\documentclass[table]{article} 
\pdfoutput=1
\usepackage{iclr2024_conference,times}

\usepackage{amsmath,amsfonts,bm}

\def\eqref#1{equation~\ref{#1}}

\def\1{\bm{1}}

\DeclareMathAlphabet{\mathsfit}{\encodingdefault}{\sfdefault}{m}{sl}
\SetMathAlphabet{\mathsfit}{bold}{\encodingdefault}{\sfdefault}{bx}{n}

\usepackage[utf8]{inputenc}
\usepackage{CJKutf8}
\usepackage[T1]{fontenc}    
\usepackage{hyperref}       
\usepackage{url}            
\usepackage{booktabs}       
\usepackage{amsfonts}       
\usepackage{nicefrac}       
\usepackage{microtype}      
\usepackage[OT1]{fontenc}
\usepackage{algorithm}
\usepackage{algorithmic}
\usepackage{xcolor}         
\usepackage{graphicx} 
\usepackage{multirow}
\usepackage{enumitem}
\usepackage{flafter}
\usepackage{listings}
\usepackage{adjustbox}
\usepackage{threeparttable}
\usepackage{mathptmx}
\usepackage{amsmath,amsthm,amssymb}
\usepackage{mathrsfs}
\usepackage{makecell}
\DeclareMathAlphabet{\mathcal}{OMS}{cmsy}{m}{n}
\title{Sequential Condition Evolved Interaction Knowledge Graph for Traditional Chinese Medicine Recommendation}

\author{
    {Jingjin Liu$^{1}$ \quad Hankz Hankui Zhu$^{1}$ \quad Kebing Jin$^{2}$ \quad Jiamin Yuan$^{3}$ \quad Zhimin Yang$^{3}$} \quad  Zhengan Yao$^{1}$\\
$^1$ Sun Yat-sen University, Guangzhou, China \\
$^2$ GuiZhou University, Guiyang, China \\
$^3$ The Second Affiliated Hospital of Guangzhou University of Traditional Chinese, Guangzhou, China \\
\texttt{liujj263@mail2.sysu.edu.cn,~zhuohank@mail.sysu.edu.cn} \\
\texttt{kbjin@gzu.edu.cn,~haopeng@illinois.edu} \\
\texttt{13580355694@126.com,~yangyovip@126.com} \\
\texttt{mcsyao@mail.sysu.edu.cn}
}

\newcommand{\red}[1] {\textcolor{red}{#1}}
\newcommand{\blue}[1] {\textcolor{blue}{#1}}
\newcommand{\ignore}[1]{{}}

\iclrfinalcopy 

\begin{document}
\begin{CJK}{UTF8}{gbsn}
\maketitle

\begin{abstract}
Traditional Chinese Medicine (TCM) has a rich history of utilizing natural herbs to treat a diversity of illnesses. In practice, TCM diagnosis and treatment are highly personalized and organically holistic, requiring comprehensive consideration of patients' states and symptoms over time. However, existing TCM recommendation approaches overlook the changes in patients' states and only explore potential patterns between symptoms and prescriptions. In this paper, we propose a novel Sequential Condition Evolved Interaction Knowledge Graph (SCEIKG), a framework that treats the model as a sequential prescription-making problem by considering the dynamics of patients' conditions across multiple diagnoses. In addition, we incorporate an interaction knowledge graph to enhance the accuracy of recommendations by considering the interactions between different herbs and patients' conditions. Experimental results on the real-world dataset demonstrate that our approach outperforms existing TCM recommendation methods, achieving state-of-the-art performance.
\end{abstract}

\section{Introduction}
\label{sec:intro}
Traditional Chinese Medicine (TCM) is an ancient and comprehensive system that has been integral to Chinese society for millennia \citep{cheung2011tcm}. TCM differs from Western medicine in light of its unique theoretical foundation, diagnosis methods, and treatment approaches, emphasizing the harmonious functioning of the body's structures \citep{zhang2015effect}. Chinese Herbal Medicine, a key component of TCM, has gained global recognition for its positive impact on various illnesses. As a result, TCM recommendation systems, which assist physicians in making informed decisions about prescribing herbs, have emerged as crucial tools. However, TCM practitioners traditionally employ observation, listening, questioning, and pulse-taking methods to understand the overall disease conditions of patients, rather than treating individual symptoms. Furthermore, TCM diagnosis and treatment prescriptions are often based on clinical experience, lacking standardization in sophisticated TCM knowledge. It is, however, essential to note that systems are not intended to replace the expertise of physicians, but rather augment it.

\begin{figure}[t]
	\centering
	\includegraphics[width=0.99\linewidth]{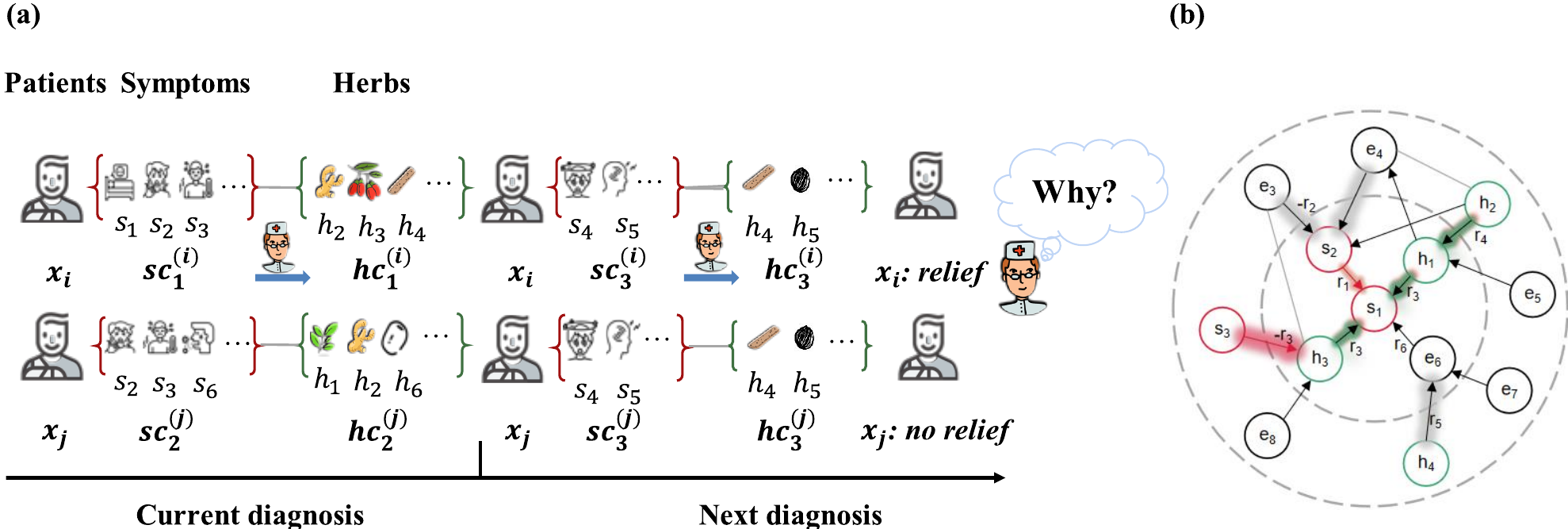}
	\caption{(a) An example of prescribing herbs based on evolution in patient symptoms; (b) An example of IKG containing information about multiple entities.}
	\label{fig:figure1}
\end{figure}

Recently, there have been approaches that have acquired better results. However, we found that there are still two shortcomings: (1) \textbf{many approaches} \citep{ruan2019discovering,jin2020syndrome,jin2021kg,yang2022multi} \textbf{primarily focus on patient symptoms or herbs, neglecting the explicit prediction of how a patient's state may change after taking medication.} \emph{As an example, consider two patients, $x_i$ and $x_j$ (as shown in Fig.\ref{fig:figure1}a), both struggling with insomnia, but with different sets of symptoms. Patient $x_i$ presents $sc_1^{(i)}=$ $\{$wakefulness, irritability, bitter mouth$\}$, while patient $x_j$ has $sc_2^{(j)}=$ $\{$dreamy, palpitations, fatigue$\}$. Subsequently, both patients took the corresponding herbal prescriptions $hc_1^{(i)}$ and $hc_2^{(j)}$, and the same symptoms set, $sc_3$, appeared at their next diagnosis. Based on the same set of symptoms, the doctor writes the same prescription. However, after the current diagnosis, patient $x_i$ experiences remission, while patient $x_j$ does not. Why is that? The answer may lie in the fact that both patients are in different states —state $o_1$ and state $o_2$ — with the same prescription $hc_3$ not accounting for these variations, potentially undermining the effectiveness of treatment.} While some Western medicine recommendation methods \citep{yang2021safedrug,shang2019gamenet,yang2022multi} consider historical data, they do not explicitly predict the patient's post-medication state. (2) \textbf{Insufficient utilization of domain knowledge.} Most methods \citep{wang2019knowledge} typically focus on mining the symptoms and prescriptions within the dataset or incorporate domain knowledge as pre-trained model inputs. However, actual TCM treatment involves four intricate steps laden with profound knowledge. Consequently, relying solely on dataset information falls short of unveiling the complexity of symptom interaction. Additionally, the lack of standardized practices in TCM makes it challenging, and many methods prescribe a fixed set of remedies, which may not be suitable for a patient's condition. \emph{For instance, if a patient describes symptoms such as $\{$headache, runny nose, cough$\}$ relying solely on current symptoms provides incomplete information. In reality, these symptoms may also be correlated with other conditions like a sore throat. Hence, depending solely on symptoms from the dataset cannot capture crucial high-level insights. To formulate appropriate herbal prescriptions, richer information is needed, considering the complex associations between symptoms as well as the compatibility between different herbs.} In this way, we can better understand the patient's condition and provide more accurate herbal treatment recommendations.

Motivated by the aforementioned shortcomings, we introduce a novel conceptual framework SCEIKG, which aims to enhance the accuracy of prescribing rational treatments by learning how patients' conditions evolve over multiple sequential diagnoses. Our approach builds upon two key observations: (1) \textbf{explicitly leverage on the change in the state of the patient after taking the medication.} We argue that this crucially hinges on the explicit as well as implicit overall condition patient's symptoms described as to why a particular relevant herbal score is coupled to a particular patient. Because each patient has a unique constitution, even when given the same prescription, the resulting changes in their condition can vary widely. Therefore, TCM recommendations must take into account the evolution of the patient's condition. To address this, we introduce a module that predicts how a patient's condition will change after taking medication. This predictive capability enables our model to make reasonable TCM recommendations, even when information about the patient's future state is unavailable. (2) \textbf{incorporating domain knowledge for symptom richness and herb compatibility.} We recognize the importance of domain knowledge in ensuring the richness of symptoms and compatibility of herbs. 
\emph{Based on the example of a patient's consecutive diagnoses, who was suffering from sleepness, bitter mouth, dry throat, etc., we leverage TCM knowledge graph domain knowledge to make extrapolations based on incomplete symptom information. By employing a GNN with IKG as additional auxiliary information, we identify that a specific herb set, including salvia miltiorrhiza and ostrea gigas, can effectively address the symptoms set. This conclusion is drawn from the long-range connections in the graph $s_{1} \stackrel{r_{1}}{\longrightarrow} s_{2} \stackrel{-r_{2}}{\longrightarrow} e_{3}  \stackrel{r_{3}}{\longrightarrow} e_{4} \stackrel{r_{4}}{\longrightarrow}\left\{h_{1}, h_{2},h_{3},...\right\}$. Further, we aggregate high-order similarity relationships and interactions among triplets using graph-based methods, enhancing our understanding of the complex relationship between herbs and symptoms (as depicted in Fig.\ref{fig:figure1}b).} Inspired by \citep{wang2019kgat} and \citep{tu2021conditional}, a hybrid structure, the Interaction Knowledge Graph (IKG), which combines the knowledge graph neighborhood knowledge of TCM and the symptoms-herbs graph to model the intricate relationships between symptoms and herbs. Also, we employ a strategy that involves training both IKG and sequential recommendation models to seamlessly integrate structured and unstructured information. This integrated approach provides a more comprehensive understanding and prediction of real-world scenarios, empowering our recommendation model with dynamic capabilities. Meanwhile, we update the graph structure based on the correlation-based attention mechanism employing the domain knowledge of IKG, which is accomplished by propagating different relation types among entities in the IKG, thus alleviating the issue of herb compatibility to some extent.

We end with a thorough empirical evaluation of our approach to our new collection of real-world data, where we explore the benefits of assessing the condition of the patient after taking medicine. Our results show that learning in a way that accounts for patients' symptoms set and the change of conditions by the sequential diagnoses has significant advantages on TCM recommendation tasks. 

\begin{figure}[t]
	\centering
	\includegraphics[width=0.99\linewidth]{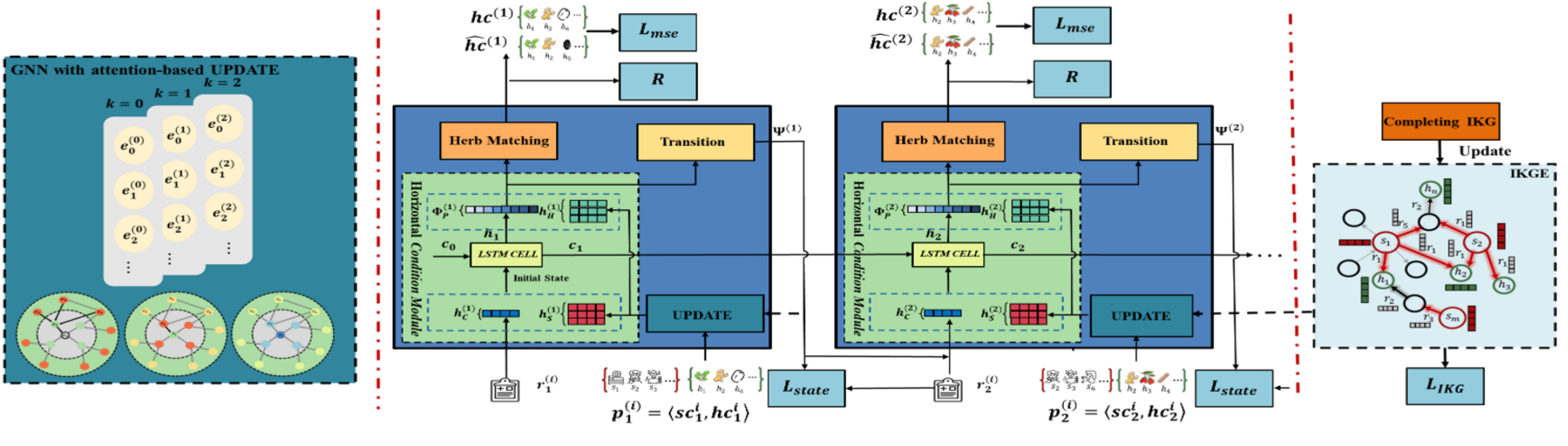}
	\caption{Schematic illustration of the proposed method.}
	\label{fig:figure2}
\end{figure}

\section{Related Work}
\label{sec:related}
\textbf{Traditional Recommendations.} Graph neural networks (GNNs) \citep{wu2020comprehensive,scarselli2008graph,kipf2016semi} have proven to be highly effective in various applications and are gaining popularity in traditional recommender systems. GCN-based embedding approaches \citep{wang2019neural,wang2018ripplenet,wang2019kgat,tu2021conditional} have been proposed to capture collaborative signals and propagate user preferences. Apart from that, \citep{sun2019bert4rec} and \citep{feder2022eye}, have also been proposed to model user behavior sequences and improve relevance prediction robustness. Traditional recommendation methods \citep{wang2019kgat,zhang2020reward, Zou2022ImprovingKR} only predict the relevance of items to a single user and do not optimize their relationships. Therefore, TCM recommendation methods need to jointly consider the relationship between the symptom set and herb set. 

\textbf{TCM Recommendations.} TCM recommendations can be roughly categorized into topic models and graph-based models. Topic models \citep{yao2018topic,chen2018heterogeneous,wang2019knowledge,ma2017discovering} were applied to mine similarities between co-occurring symptoms and herb words under the same topic, and introduce TCM domain knowledge into topic models to capture herb compatibility regularities. However, those topic models cannot analyze complex interactions. Recently, graph-based models \citep{ruan2019discovering,jin2020syndrome,yang2022multi} have been used extensively in TCM recommendations. And \citep{zhang2023knowledge} have built small-scale knowledge graphs to train recommendation models. As most of this literature deals with mining potential relationships between symptoms and herbs, there is no work, as far as we know, that modeled the condition transition in TCM recommendations. Moreover, we first propose the use of a sequential follow-up dataset to model the impact of the condition evaluation before and after medication on TCM recommendations. Additionally, We also constructed a large-scale knowledge graph with different relations and combined it with the dataset to train the model together.
\citep{ruan2019discovering} use auto-encoders with meta-path to discover TCM information, \citep{jin2020syndrome} proposed a syndrome-aware mechanism to acquire the embedding of symptom sets in the prescription through graph convolution networks (GCNs). \citep{yang2022multi} also integrates herb properties through multi-layer information fusion with GCN.

\section{Problem Formulation}
\label{sec:setting}
We denote the set of symptoms by $\boldsymbol{S}=\{s_1,s_2,...,s_M\}$ and the set of herbs by $\boldsymbol{H}=\{h_1,h_2,...,h_N\}$, respectively. Note that a symptom $s_i$ is represented by a TCM symptom term, e.g., \emph{抑郁症(depression)}; a herb $h_i$ is represented by a TCM herb term, e.g., \emph{茯苓(tuckahoe)}. We define an IKG by $\mathcal{G}=(\mathcal{E},\mathcal{R},\mathcal{T}, \mathcal{A})$, where $\mathcal{E}$ is a set of \emph{entities} and $\mathcal{R}$ is a set of \emph{relations}. $\mathcal{T}$ is a set of triples $\mathcal{T}=\{(h,r,t)|h \in \mathcal{E}, t\in \mathcal{E}, r\in \mathcal{R}\}$, where each triples means there is a relation $r$ from \emph{head} entity $h$ to \emph{tail} entity $t$. Specifically, $\mathcal{E}$ consists of symptoms $\boldsymbol{S}$, herbs $\boldsymbol{H}$, and other entities such as \emph{pharmacology}, \emph{efficacy}, \emph{diseases}, \emph{examination} and \emph{diagnosis}, which were extracted from TCM datasets \citep{yao2018topic} to help entail relations between symptoms and herbs directly or indirectly (c.f. Appendix \ref{sec:ZzzTCM Dataset and IKG Processing}). A relation $r \in \mathcal{R}$ indicates the relationship among entities, e.g., 
\emph{symptoms-related herbs}.  
The adjacency matrix $\mathcal{A}=[a_{e_i,e_j}]_{V\times V}$ was built based on different types of edge relationships by the co-occurring probabilities using Normalized Pointwise Mutual Information \citep{bouma2009normalized}:
\begin{equation}
\label{eq: IKG}
a_{e_i,e_j} =\left\{\begin{array}{ll}
1-\frac{log(p(e_i)p(e_j))}{log p_r(e_i,e_j)}, & \text { if }\left(e_{i}, e_{j}\right) \  \text{co-occur \ in \ $\mathcal{T}$ }  \\
0, & \text { otherwise }
\end{array}\right.
\end{equation}
where $p_r(e_i,e_j)$ is the joint probability of $e_i$ and $e_j$ with relation $r$, and $p(e_i)$ (or $p(e_j)$) is the probability of occurrences of $e_i$ (or $e_j$) in all relations. $V$ is the number of entities in $\mathcal{G}$.


In this paper, we denote the set of patients by $\boldsymbol{X}=\{x_1,x_2,x_3,...,x_{J}\}$, and the set of sequences of \emph{diagnoses} by $\Omega=\{\omega_j|\omega_j=\langle \omega_j^{(1)},\omega_j^{(2)},\ldots, \omega_j^{(T_j)}\rangle, 1\leq j\leq J\}$, where 
$\omega_j^{(t)} = (O^{(t)}_{j}, \mathbb{S}^{(t)}_{j}, \mathbb{H}^{(t)}_{j})$ is the $t$-th diagnosis for $1\leq t\leq T_j\}$, and $T_j$ is the number of diagnoses of patient $x_j$. $O_j^{(t)}$ is the description of patient $x_j$ during the $t$-th diagnosis in the form of natural language text, $\mathbb{S}^{(t)}_{j}\subseteq \boldsymbol{S}$ is a set of symptoms of patient $x_j$ in $t$-th diagnosis, and $\mathbb{H}^{(t)}_{j}\subseteq \boldsymbol{H}$ is a set of herbs given to patient $x_j$ in $t$-th diagnosis. Our TCM recommendation problem can be formulated by: given a set of sequences of diagnoses $\Omega$ and an initial IKG $\mathcal{G}$, we aim to learn a model $\mathcal{M}_\Theta$ to recommend a set of herbs to a new patient $x_{new}$ in the $k$-th diagnosis based on the patient's historical sequence of diagnoses $\omega_{new}=\langle \omega_{new}^{(1)},\omega_{new}^{(2)},\ldots, \omega_{new}^{(k-1)}\rangle$, the current description $O_{new}^{(k)}$ and the current symptom $\mathbb{S}^{(k)}_{new}$, i.e., 
\[\mathbb{H}^{(k)}_{new}=\mathcal{M}_\Theta(\omega_{new},O_{new}^{(k)},\mathbb{S}_{new}^{(k)}),\]
where $\Theta$ is the parameters of the model to be learned from $\Omega$ and $\mathcal{G}$. $\omega_{new}$ denotes sequential information about the patient and $\omega_{new}^{(k-1)}$ denotes the single diagnosis. Note that $\Theta$ includes both the neural network parameters and the representation parameters of entities and edges in $\mathcal{G}$.


\section{Modeling Approach}
\label{sec:approach}

In this section, we present SCEIKG, a Sequential Condition Evolved based on an Interaction Knowledge Graph learning model to enhance the accuracy of TCM recommendation. The framework, depicted in Fig.\ref{fig:figure2}, consists of three modules. (1) A heterogeneous Graph Neural Network (GNN) utilizes a hierarchical attention message passing layer and knowledge graph embedding layer to obtain embeddings for all entities. (2) A horizontal condition module learns the patient's current representation from historical records, and then takes the patient's representation as input and outputs an herbal vector that measures the similarity between each herbal representation and the patient representation. (3) A transition condition module observes the patient's progression after taking the recommended herbs, with the evolved status serving as an auxiliary indicator for upcoming diagnoses. The framework is trained with a joint objective function to ensure accurate TCM recommendations.

\subsection{Heterogeneous GNN with attention-based UPDATE}

The heterogeneous Graph Neural Network (GNN) captures information about entities through recursive propagation to update its representation. The $\mathrm{AGGREGATE}^{(k)}(\cdot)$ function integrates the entity $h$ feature from its neighboring entities $t$ conditioned to relation $r$. This operation is represented as $\alpha_h^{(k)} = \mathrm{AGGREGATE}^{(k)} \left( e_{h}^{(k-1)}, \sum_{t \in \mathcal{N}(h,r)}w_{(h,r,t)}e_{t}^{(k-1)}\right)$, where $e_h^{(k)}\in \mathbb{R}^{D_k}$ is the feature embedding of entity $h$ at layer $k$, and $\mathcal{N}(h,r)$ indicates the neighbors connected to $h$ with relation $r$. The function then propagates the integrated information to update the entity features at the next layer. To capture higher-order similarities between entities, the interaction knowledge graph (IKG) is utilized. Inspired by KGAT \citep{wang2019kgat}, the $\mathrm{UPDATE}(\cdot)$ function updates the weights $\mathbb{W}$ of relations in the IKG, indicating the information propagation strength from $t$ to $h$ based on the relationship $r$. The weight value $w_{(h,r,t)}=\mathrm{UPDATE}\left(\{w_{(h,r,t)}\;|\; (h,r,t) \in \mathcal{G}\}\right) = Softmax\left(\left(W_re_h+e_r\right)*W_re_t \right)$ is calculated using an attention mechanism, considering the correlation between $e_h \in \mathbb{R}^{D_{in}}$ and $e_t \in \mathbb{R}^{D_{in}}$ in the specific-relation $r$ space. The weight value depends on the transformation matrix $W_r$ of relation $r$. 
The final weight matrix $\mathbb{W} \in \mathbb{R}^{V \times V}$ is obtained from the $\mathcal{A}$ by graph-based Laplcian \citep{kipf2016semi} calculation to assess the connections between all entities. And the $\mathrm{AGGREGATE}^{(k)}(\cdot)$ and $\mathrm{COMBINE}^{(k)}(\cdot)$ can formulate in the matrix as follows:

\begin{equation}
\label{eq: equation2}
E^{(k)} = SUM\left(\boldsymbol{NN_1}\left( \left(E^{(k-1)}+\mathbb{W}E^{(k-1)}W_1^{(k)}\right); W_3^{(k)}\right), \boldsymbol{NN_2}\left(\left( E^{(k-1)} \odot \mathbb{W}E^{(k-1)}W_2^{(k)}\right); W_4^{(k)} \right) \right)
\end{equation}

where $E^{(k)} = [e_1^{(k)},...,e_h^{(k)}] \in \mathbb{R}^{V \times D_{k+1}}$ is the stack of entity feature vectors, and $e^{(0)} \in \mathbb{R}^{D_{in}}$ initialization using a uniform distribution. $\boldsymbol{NN_1}(\cdot)$ and $\boldsymbol{NN_2}(\cdot)$ is the forward propagation neural network with an activation function, $W \in \mathbb{R}^{D_{k}\times D_{k+1}}$ is the network weights. The final entity representations $E= Concatenate\left(E^{(0)},.., E^{(k)}\right) \in \mathbb{R}^{V \times D_{out}}$ is defined simply as concatenating the entity features of all layers, where the $D_{out}$ is the dimension of the embedding space.

Knowledge graph representation is an effective way to enhance the completeness of the links between entities and enable the provision of more nuanced and sophisticated information. Meanwhile, we also learn the representation of entities using the TransR \citep{lin2015learning} combined with RotatE \citep{sun2018rotate} to make an entity play different roles in different tripels to complement the links of the $\mathcal{G}$:

\begin{equation}
\label{eq: equation3}
\boldsymbol{f} \left(h,r,t\right) = C * ||Sin(W_re_h+e_r-W_re_t)||_1
\end{equation}

$W_r \in \mathbb{R}^{R \times D_{in}} $ is the transformation matrix of relation $r$ and $C$ is the modulus of constraint, $||\cdot||_1$ is the $L_1$-norm. A lower score of $\boldsymbol{f}(h, r, t)$ indicates the triplet is more likely to be true and vice versa. By completing the links between entities, we further update the entity representation $E \in \mathbb{R}^{V \times D_{out}}$.


\subsection{Horizontal Condition Module}
The TCM emphasizes the importance of maintaining a harmonious body structure, as well as considering the patient's overall well-being. 
Therefore, it is crucial to gain a comprehensive understanding of the patient's core health state. 

\textbf{Condition Representation.} To extract the patient's state, we employ Bidirectional Encoder Representations from Transformers (BERT) \citep{devlin2018bert} pre-trained transformer base model. With the exception of fine-tuning transformer models, the condition representation $h_\mathbb{C}^{(t)} \in \mathbb{R}^{l}$ is not only represented by the $l$ dimensional hidden state $h_{bert}^{(t)} $ of the "[CLS]" token in the last layer but is also fed into average pooling layer $g(\cdot)$ that extracted the overall patient condition vector by assigning weights, as well as focusing on critical and effective information. The process of condition representation can be formulated $h_\mathbb{C}^{(t)} = \sum_i^{\Gamma}g(h_{bert}^{(t)}; W_5)_i h_{i,bert}^{(t)}$, where $\Gamma$ is the length of the record sequence $O^{(t)}$, the average pooling layer $g(\cdot): \mathbb{R}^{\Gamma \times l} \rightarrow \mathbb{R}^{l} $ is combined the $h_{bert}^{(t)} \in \mathbb{R}^{\Gamma \times l}$ with assigned attention weight between word $i$ and $j$ to get the condition representation $h_\mathbb{C}^{(t)} \in \mathbb{R}^{l}$. And we apply a feed-forward neural network, $\boldsymbol{NN_3}(\cdot):\mathbb{R}^{l} \rightarrow \mathbb{R}^{D_{out}}$, for dimensionality transformation. To avoid over-fitting, we also apply a high rate of dropout to this high-dimensional condition representation.

\textbf{Symptoms Representation.} 
In TCM recommendation, instead of modeling relationships between single users and single items, sets of symptoms and sets of herbs are considered. Encoding the symptoms set $\mathbb{S}^{(t)}$ to the multi-hot symptoms $sc^{(t)} \in {\{0,1\}}^{M}$ and the shared symptoms embedding table $E_s \in E: \mathbb{R}^{M \times D_{out}}$ that explicitly aggregating the multi-hop connectivity information to related symptoms, herbs and the similar entity representation in the knowledge graph. We introduce the corresponding symptoms into the embedding space by vector-matrix dot product, represented as $ h^{(t)}_{\mathbb{S}} = \sum_{i:sc^{(t)}_{i}=1}^{M} sc_{i}^{(t)}E_{s,i}$, where $h^{(t)}_{\mathbb{S}} \in \mathbb{R}^{D_{out}}$ stores the embedding vector for particular symptoms in the $t$-th diagnosis symptoms for one patient.

\textbf{Horizontal Patient Representation.} It is possible that a health snapshot will not be sufficient to make treatment decisions. For example, at the previous ${(t-1)}$-th diagnosis, the patient had insomnia, and the doctor prescribed herbs that provided some relief, which observed the patient in a new state $\Psi^{(t-1)}$. However, at the current $t$-th diagnosis, the patient did not mention the insomnia-related symptoms, but only presented with a headache. Encoding the herbs set $\mathbb{H}^{(t)}$ to the multi-hot herbs $hc^{(t)} \in {\{0,1\}}^{N}$. Here, we use an LSTM \citep{hochreiter1997long} model to dynamically model the patient's historical states $\Psi^{(t)} = [\Psi^{(t-1)}, \Psi^{(t-2)},...,h_C^{(0)}]$ and eventually obtain a comprehensive patients representation $\Phi_P^{(t)}$:


\begin{equation}
\label{eq: patient embedding3}
\Phi^{(t)}_P = \boldsymbol{LSTM} \left( \left(  \Pi \left(h_\mathbb{S}^{(t)},\Psi^{(t-1)}\right) ,C^{(t-1)} \right),...,\left( \Pi \left(h_\mathbb{S}^{(0)},h_\mathbb{C}^{(0)}\right) ,C^{(0)}\right); \; W_6\right)
\end{equation}

where the $C^{(t-1)}$ includes the hidden state $h^{(t-1)} \in \mathbb{R}^{hidden\_dim}$ and cell state $c^{(t-1)} \in \mathbb{R}^{hidden\_dim}$, and the initial state $h^{(0)}$ and $c^{(0)}$ are all-zero vectors, $C$ will be passed down. The State $\Psi^{(t-1)}=\boldsymbol{T}\left(\Phi^{(t-1)}_P,hc^{(t-1)}\right)$ is obtained by transferring the state $\Phi^{(t-1)}_P$ at the $(t-1)$-th diagnosis to the state after taking the herbs $hc^{(t-1)} \in \{0,1\}^N$.  A more compact representation $\Pi: \mathbb{R}^{2D_{out}}\rightarrow \mathbb{R}^{D_{out}}$ of the patient is created by concatenating the historical condition representation $\Psi^{(t-1)} \in \mathbb{R}^{D_{out}}$ and symptoms representation $h_\mathbb{S}^{(t)} \in \mathbb{R}^{D_{out}}$ as a double-long vector, along with a layer of self-attention. Additionally, The operation of transition condition $\boldsymbol{T}(\cdot)$ will be introduced in section \ref{sec:Transition Condition Module}.

\textbf{Patient-to-herb Matching.}  
After obtaining the patient's horizontal representation $\Phi_{P}^{(t)} \in \mathbb{R}^{D_{out}}$, we aim to identify the most relevant herbs from the herbs embedding table $E_h \in E: \mathbb{R}^{N\times D_{out}}$. To achieve this, we perform an inner product to calculate the scores between $E_h$ and $\Phi_{P}^{(t)}$, followed by the application of the sigmoid function $\sigma(\cdot)$ to complete the operation $\boldsymbol{P}(\cdot)$ of herbal recommendation. The process of operation is $\hat{Y}^{(t)}= \boldsymbol{P}\left(\Phi_{P}^{(t)}, E_h\right) = \sigma\left(\Phi_{P}^{(t)}E_h^{T}\right)$, where $\hat{Y}^{(t)} \in \mathbb{R}^{N}$ of every element stores a matching score for one herb. Finally, we obtain the recommended herbal set $\mathbb{H}^{(t)}$ based on $\hat{Y}^{(t)}$.

\subsection{Transition Condition Module}
\label{sec:Transition Condition Module}
In practice, treating a disease is a complex and gradual process, and it is often difficult for patients to achieve complete recovery with a single treatment. Since each patient's status changes differently, representing their post-herb status implicitly is essential. The operation of transition condition $\boldsymbol{T}(\cdot)$ is designed to model the patient's condition shift after taking the herbs $hc^{(t-1)} \in \{0,1\}^N$. Specifically, we obtain the one-dimensional convolution results of the herb representations $h_\mathbb{H}^{(t)} = \boldsymbol{Conv1D}\left(\boldsymbol{P}\left(\Phi_{P}^{(t)}, E_h\right)E_h; \; W_7\right)$ to capture the global information and eliminate the position effect. Inspired by \citep{liu2018deep}, herb and patient interactions $h^{(t)}_\mathbb{I}$ are considered by multiplying the matrix elements of the herb representation $h_\mathbb{H}^{(t)} \in \mathbb{R}^{D_{out}}$ and the horizontal patient representation $\Phi_{P}^{(t)} \in \mathbb{R}^{D_{out}}$. Finally, the one-dimensional convolution results of the herb representation $h_\mathbb{H}^{(t)}$, the interaction representation $h^{(t)}_\mathbb{I} \in \mathbb{R}^{D_{out}}$ and the patient representation $\Phi_{P}^{(t)}$ are concatenated into an embedding space to represent the transition conditional representation $\Psi^{(t)}$ of the patient after taking the herbs thus achieving a state transfer. Formally, the transition condition module $\boldsymbol{T}(\cdot)$ can be expressed as:

\begin{equation}
\label{eq: symptoms embedding}
\Psi^{(t)} = \boldsymbol{T} \left(\Phi^{(t)}_P,\boldsymbol{P}\left(\Phi^{(t)}_P, E_h\right)\right) = \boldsymbol{NN_4} \left(Concatenate\left({\Phi_{P}^{(t)},(\Phi_{P}^{(t)}\odot  h_\mathbb{H}^{(t)}),h_\mathbb{H}^{(t)}}\right);\; W_8\right)
\end{equation}

where the $\odot$ is the Hadamard product. Note that, the transition condition representation $\Psi^{(t)} \in \mathbb{R}^{D_{out}}$ is represented as the condition representation $h_\mathbb{C}^{(t+1)}$ at the next diagnosis. $\boldsymbol{NN_4}(\cdot)$ represents a feed-forward neural network, and its primary function is to perform nonlinear mapping on input data. $W_8$ is the weight parameter of the neural network.

\subsection{Model Training with Objective Function}
Our approach to robust learning is based on regularized risk minimization, where regularization acts to discourage the appearance of two mutually exclusive herbs in the recommendation. Our joint objective is:

\begin{equation}
\label{eq: equation6}
\mathop{argmin}\limits_{\boldsymbol{T},\; \boldsymbol{P},\; \Theta } L_{mse}\left(\boldsymbol{P}^{(t)},\Theta \right)  + L_{state}\left(\boldsymbol{T}^{(t)};\Theta \right) + \lambda \boldsymbol{R} \left(\boldsymbol{P}^{(t)}, \mathbb{W}, \Theta \right) + L_{IKG}\left( \mathcal{G},\Theta \right) + \lambda_{\Theta}||\Theta||_2^2
\end{equation}

where $\boldsymbol{P}$ is the function prediction herbs, $L_{mse}=\sum\limits_{i=1}^{N}(hc_i^{(t)}-\hat{Y}_i^{(t)})^2$ is the average loss w.r.t an empirical MSE loss function. The objective involves evaluating the distance between the recommended herbs set and the ground truth herbs set. And $\boldsymbol{T}^{(t)}$ is a function of transition condition and $L_{state}=Cos(h_\mathbb{C}^{(t+1)},\Psi^{(t)}) = \frac{h_\mathbb{C}^{(t+1)} \cdot \Psi^{(t)}}{||h_\mathbb{C}^{(t+1)}||*||\Psi^{(t)}||}$ is minimized the state $\Psi^{(t)}$ after taking the medication and the next state $\Phi_P^{(t+1)}$. The regularization scheme $\boldsymbol{R}=-\sum_{i=1}^{N}\sum_{j=1}^{N}\mathbb{W}_{ij}\hat{Y}^{(t)}_i \hat{Y}^{(t)}_j$ is to penalize the $\boldsymbol{P}^{(t)}$ for violating certain pair of herbs, where the $\mathbb{W}_{i,j}$ from the weight $\mathbb{W}$ in $\mathcal{G}$ indicates the strength of compatibility between $i$-th herb and $j$-th herb. If they are mutually exclusive, then $\mathbb{W}_{i,j}^{h}=0$. Knowing the complete TCM graph of all herbs (e.g. the degree to which herbs are repulsive or interact with each other) can certainly help, but reliance on mass is impractical. 

As we all know, constructing a complete TCM knowledge graph is a difficult task that relies on extensive data support. Therefore, the loss $L_{IKG}=\sum\limits_{\left(h,r,t,t'\right) \in \mathcal{T}} -ln\sigma\left(\boldsymbol{f}\left(h,r,t'\right)-\boldsymbol{f}\left(h,r,t\right)\right)$ is to complete the TCM knowledge graph, allowing for the inference of useful information that was not initially available. The $\mathcal{T} = \{(h,r,t,t')|(h,r,t)\in \mathcal{G},(h,r,t,t')|(h,r,t) \notin \mathcal{G}\}$ is the broken triplet constructed by replacing one entity in a valid triplet randomly. Also, $\lambda_\Theta$ controls the $L_2$ regularization strength to prevent over-fitting. Note that the above loss functions are defined for a single diagnosis. During the training, loss backpropagation will be conducted at the patient level by the averaged losses across all diagnoses.
In section \ref{sec:exp} we will see that these can go a long way toward learning robust models in practice and The detailed algorithmic will be further presented in Appendix \ref{sec: Training Algorithm for SCEIKG model and Inference}.

\begin{table}[t]
    \centering
    \caption{Performance Comparison on ZzzTCM Dataset.}
    \label{tab:results}
    \resizebox{\columnwidth}{!}{
	\begin{tabular}{c c c c c c c c c c} 
	      \toprule
            \multirow{2}{*}{\textbf{Models}} & \multicolumn{3}{c}{\textbf{Precision}} & \multicolumn{3}{c}{\textbf{Recall}} & \multicolumn{3}{c}{\textbf{F1}}\\ 
            \cmidrule(r){2-4} \cmidrule(r){5-7} \cmidrule(r){8-10}
			& P@5 & P@10 & P@20 & R@5 & R@10 & R@20 & F1@5 & F1@10 & F1@20   \\
            \midrule
			BPR      & $0.4087$ & $0.3418$ & $0.2563$ & $0.2066$ & $0.3384$ & $0.5004$ & $0.2669$ & $0.3298$ & $0.3300$\\
            GCN      & $0.4765$ & $0.3711$ & $0.2792$ & $0.2287$ & $0.3536$ & $0.5557$ & $0.3017$ & $0.3599$ & $0.3613$\\
			KGAT     & $0.4832$ & $0.3852$ & $0.2956$ & $0.2434$ & $0.3835$ & $0.5822$ & $0.3152$ & $0.3730$ & $0.3812$\\
            GAMENet & $0.5066$ & $0.4176$ & $\textbf{0.3096}$ & $0.2557$ & $0.4151$ & $\textbf{0.6027}$ & $0.3300$ & $0.4037$ & $\textbf{0.3976}$\\
            SafeDrug & $0.5038$ & $0.4082$ & $0.3000$ & $0.2562$ & $0.4105$ & $0.5926$ & $0.2672$ & $0.3364$ & $0.3534$\\
            SMGCN    & $0.5248$ & $0.4121$ & $0.3027$ & $0.2637$ & $0.4136$ & $0.5900$ & $0.3380$ & $0.3982$ & $0.3887$\\
            KDHR     & $0.4329$ & $0.3787$ & $0.2872$ & $0.2229$ & $0.3862$ & $0.5689$ & $0.2680$ & $0.3710$ & $0.3715$\\
            \textbf{Ours}     & $\textbf{0.5477}$ & $\textbf{0.4275}$ & $0.3087$ & $\textbf{0.2727}$ & $\textbf{0.4243}$ & $0.6010$ & $\textbf{0.3538}$ & $\textbf{0.4128}$ & $0.3973$ \\
			\bottomrule
		\end{tabular}}
\end{table}
\begin{table}[t]
    \centering
    \caption{The difference and intersection herbs prescribed by our model and TCM doctor according to clinical symptoms and records of the same patient for two diagnoses.}
    \label{tab: herb example}
    \resizebox{\textwidth}{!}{
	\begin{tabular}{c l l l} 
	      \toprule
            \makecell[c]{\multirow{2}{*}{\textbf{Sequential diagnoses}}}  &\makecell[l]{\multirow{2}{*}{\textbf{Symptom Set}}} & \multicolumn{2}{c}{\textbf{Herb Set}} \\ 
            \cmidrule(r){3-4} 
			&  & \makecell[c]{\textbf{SCEIKG}} & \makecell[c]{\textbf{TCM doctor}} \\
            \midrule

             & 抑郁症(depression)  & \red{\textbf{黄芩(scutellaria baicalensis)}} & \red{\textbf{黄芩(scutellaria baicalensis)}} \\
             & 口干(xerostomia)  & \red{\textbf{炙甘草(glycyrrhiza uralensis)}} & \red{\textbf{炙甘草(glycyrrhiza uralensis)}} \\
			& 大便费力(dyschezia)  & \red{\textbf{生姜(ginger)}} & \red{\textbf{生姜(ginger)}} \\
            & 入睡困难(insomnia)  & \red{\textbf{大枣(jujube)}} & \red{\textbf{大枣(jujube)}} \\
First diagnosis   & 眠浅易醒(light sleep, easy to wake up)  & \red{\textbf{人参(ginsen)}} & \red{\textbf{人参(ginsen)}} \\
            & 乏力(fatigue)  & 桂枝(cinnamomum cassia) & 北沙参(radix adenophorae) \\
            & 胸闷(chest tightness)& 茯苓(tuckahoe) & 柴胡(bupleuri radix) \\
            & 四肢麻木(numbness of limbs)  & 白芍(paeonia lactiflora) & 天花粉(flos rosae rugosae) \\
            & 舌淡红(pale red tongue)  &牡蛎(ostrea gigas)  &  \\
            & 下睑淡白(pale lower eyelid)  & 干姜(zingiber officinale) &  \\
            \cmidrule(r){1-4}
            \multicolumn{4}{c}{\textbf{p@10=0.5000 r@10=0.6250 f1@10=0.5556}} \\
            \cmidrule(r){1-4}
            & 口干(xerostomia)  & \red{\textbf{黄芩(scutellaria baicalensis)}} & \red{\textbf{黄芩(scutellaria baicalensis)}} \\
            & 惊恐(panic)  & \red{\textbf{赤芍(paeonia lactiflora)}} & \red{\textbf{赤芍(paeonia lactiflora)}} \\
            & 焦虑 (anxiety)  & \red{\textbf{炙甘草(glycyrrhiza uralensis)}} & \red{\textbf{炙甘草(glycyrrhiza uralensis)}} \\
            & 入睡困难(insomnia)  & \red{\textbf{大枣(jujube)}} & \red{\textbf{大枣(jujube)}} \\
            & 眠浅易醒(light sleep, easy to wake up)  & \red{\textbf{生姜(ginger)}} & \red{\textbf{生姜(ginger)}} \\
            & 乏力(fatigue)  & \red{\textbf{清半夏(ternate pinellia)}} & \red{\textbf{清半夏(ternate pinellia)}} \\
Second diagnosis    &  胸闷(chest tightness)  & 茯苓(tuckahoe) &  \\
            &  四肢麻木(numbness of limbs)  & 人参(ginsen) &  \\
            &  小便频急(frequent urination)  & 桂枝(cinnamomum cassia) &  \\
            &  右手心热(palm heat)  & 炒六神曲(medicated leaven) &  \\
            &  舌淡红(pale red tongue)  &  &  \\
            &  苔薄(thin fur)  &  &  \\
            &  下睑淡白边偏红 (pale lower eyelid with reddish edges)  &  & \\
            \cmidrule(r){1-4}
            \multicolumn{4}{c}{\textbf{p@10=0.6000 r@10=1.0000 f1@10=0.7500}} \\
			\bottomrule
		\end{tabular}}
\end{table}

\begin{figure}[t]
	\centering
	\includegraphics[width=0.99\linewidth]{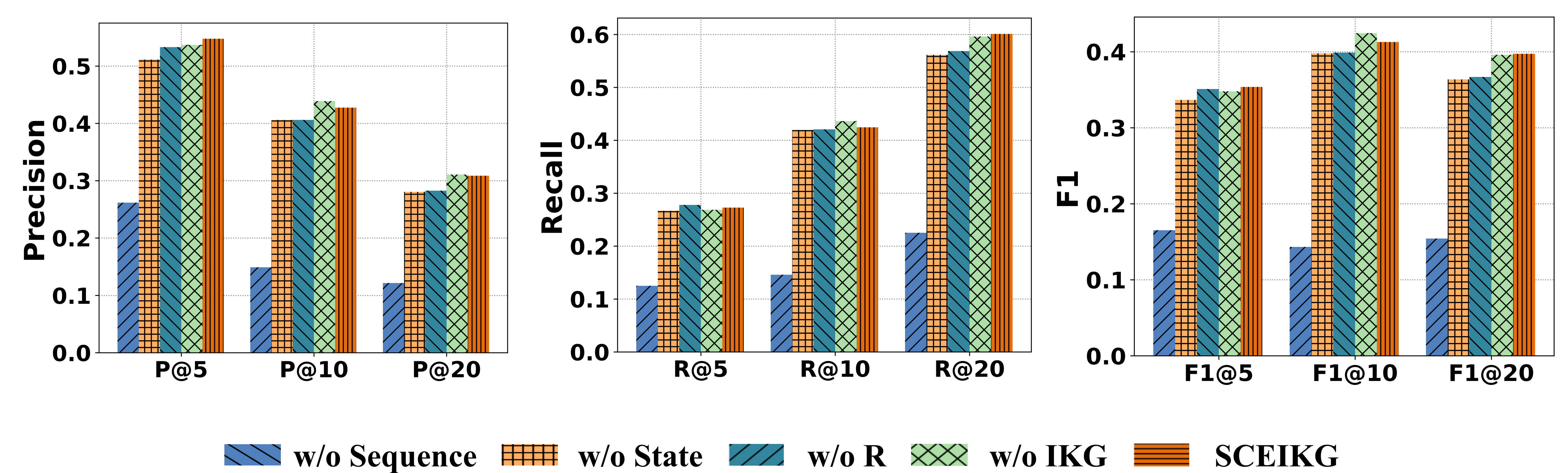}
	\caption{Performance of different variants of SCEIKG on different evaluation metrics.}
	\label{fig:ablation}
\end{figure}

\section{Experiments and Results}
\label{sec:exp}
In this section, we present the effectiveness of the model by comparing the performances of different models. Also, we conducted other experimental analyses. The Appendix includes further details on data descriptions, model architectures, training inference, experimental settings, parameter sensitivity, and the interpretative experiments of herb compatibility and embedding visualization.
\subsection{Dataset Description}
\textbf{ZzzTCM Dataset}
To guarantee the authenticity of the TCM recommendation data, we further collect a new sequential real-world dataset that includes patients' multiple diagnoses, named ZzzTCM (a play on "Zzz" for sleep and TCM for traditional Chinese medicine). We use the medical records from the Guangdong Provincial Chinese Hospital as the data source. The hospital had already sought the consent of the related patients to use their medical history for academic research. In contrast with data directly crawled from online communities, these hospital records are guaranteed to be actual cases diagnosed by doctors, which makes them of higher quality. A total of 17000 history records were provided to us, from which we selected 751 patient records and TCM prescriptions from TCM practitioners, with each patient undergoing $1\sim17$ multiple follow-up diagnoses. It is important to note that we utilized ChatGPT to help us extract the symptom set $\mathbb{S}$ from the historical records. Detailed descriptions of the dataset and part of the IKG are available in Appendix \ref{sec:ZzzTCM Dataset and IKG Processing}.

\subsection{Evaluation Metrics}
To evaluate the performance of $Top$-$K$ recommended herbs, we adopt three evaluation metrics: Precision@$Top$-$K$, Recall@$Top$-$K$, and F1@$Top$-$K$. The $Prescision$ score indicates the hit ratio of herbs is true herbs. And the $Recall$ describes the coverage of true herbs as a result of a recommendation. $F_1$ is the harmonic mean of precision and recall. In particular, we obtain the score of evaluation in the test data by taking the average of patients' diagnoses.


\subsection{Comparisons with Baselines}
We evaluate the performance of SCEIKG by comparing it against several baseline models from different methods. As illustrated in Table \ref{tab:results}, the traditional recommendation approach, \textbf{BPR} \citep{rendle2012bpr},\textbf{GCN} \citep{kipf2016semi}, \textbf{KGAT} \citep{wang2019kgat}, \textbf{GAMNet} \citep{shang2019gamenet}, \textbf{SafeDrug} \citep{yang2021safedrug}, \textbf{SMGCN} \citep{jin2020syndrome}, and \textbf{KDHR} \citep{yang2022multi}. Note that GAMENet and SafeDrug are not considered baseline since they are primarily for Western drug recommendations and require additional ontology data. Also, when applying our dataset in KDHR, we removed the module for the herb knowledge graph. In contrast, our model incorporates the condition changes, resulting in superior accuracy for TCM recommendations. As can be seen from Table \ref{tab:results}, our model outperforms GAMNet in $Top$-5 and $Top$-10 recommendations, although there is a slight difference in $Top$-20. However, in real TCM recommendations, a smaller number of recommendations presents a more significant challenge to the model.
As a result, SCEIKG demonstrates significant advancements over the baseline models, showcasing its robust predictive power in herb prediction based on multiple patient diagnoses. Our experimental setting and specific parameter settings of baseline models are provided in Appendix \ref{sec:experimental setttings}. 


\subsection{Experimental Analysis}
\label{sec:experimental analysis}
\textbf{Herb Recommendations} We conduct a case study to verify the rationality of the herb recommendation for our proposed model. Table \ref{tab: herb example} shows examples in the herb recommendations scenario. Given the symptoms set for patient $x_j$, our proposed model generates an herb set to cure the listed symptoms. In the Herb Set column, the bold red font indicates the common herbs between the herb set recommended by our model and the herbs prescribed by TCM doctors. While our model also recommended some herbs not prescribed by the doctor and there are some discrepancies between the prescribed herbal prescriptions by SCEIKG and the actual prescriptions, their appropriateness for the symptom set has been verified by the TCM doctor. The initial diagnosis's prescription was considered by the doctors to be more applicable to the given symptom set than the real prescription, thus affirming the efficacy of our model. For the subsequent diagnosis, the recommended prescription showed no significant deviations from the real prescription, which was likewise deemed suitable for treating the symptom set.

\textbf{Ablation Study} To further strengthen the credibility of our model, we conducted comprehensive comparisons with its variants to highlight the significance of each component. We introduced four model variants: (1) \textbf{SCEIKG w/o Sequence}: this variant applies the model without considering multiple diagnoses for sequential herb recommendation and the transition condition module. (2) \textbf{SCEIKG w/o State}: this variant does not take into account constraints in the patient's condition (denoted as $L_{state}$). (3) \textbf{SCEIKG w/o $\boldsymbol{R}$}: this variant excludes herbal compatibility constraints (denoted as $\boldsymbol{R}$). (4) \textbf{SCEIKG w/o IKG}: this variant is based on the initial model but excludes the TransR combined with the RotatE embedding component and the correlation attention mechanism, and we train the model without $L_{IKG}$. As shown in Fig.\ref{fig:ablation}, which can confirm the importance of each component of the model. In particular, the reason for the very poor performance of SCEIKG without sequence is that it relies on whether sequential information and transition conditions involve the model or not. We observed that considering the changes in the patient's condition after taking the herbs significantly improves the predictive performance. We note that cases with IKG are slightly weaker than cases without IKG on some of the evaluation metrics, and this result is analyzed in Appendix \ref{sec:additional case study}.



\section{Conclusion}
\label{sec:disc}
Our paper argues that to cope with an accurate herb recommendation, learning must take into account how internal changes in taking medication for patients. Toward this, we investigate the TCM recommendation task from the novel perspective of incorporating the sequential diagnoses for patients and develop a condition module to simultaneously learn the condition embedding and guide the next diagnosis.  We also introduce a knowledge graph to enhance our model. Experiments on a real-world sequential TCM dataset demonstrate the superiority of our proposed model, which mimics the consultation process of experienced doctors. As for the limitations, the patient's state transfer is implicit, and the generalization ability of the model requires further consideration. In future work, we aim to address these limitations by representing the patient's state transfer explicitly, enhancing model robustness and enriching the TCM domain-specific knowledge, including dosage information and contraindications, within the interaction knowledge graph.

\bibliographystyle{iclr2024_conference}
\bibliography{iclr2024_conference}

\newpage
\appendix


\begin{table}
    \centering
    \caption{Data Statistics}
    \label{tab:data statistics}
    \resizebox{\textwidth}{!}{
    \begin{threeparttable}
    \begin{tabular}{lll}
	      \toprule
                                    & \textbf{Items} & \textbf{Size} \\ 
            \midrule
			\multirow{3}{*}{ZzzTCM} & \# of diagnoses / \# of patients   & $ 2761 / 751$ \\
			                      & symptoms. / herbs. space size   & $ 6562 / 387$ \\
			                        & avg. / max \# of diagnoses          & $ 3.68 / 17$ \\ 
            \midrule  
			\multirow{8}{*}{IKG}    & entities   & $ 344092 $ \\
                                    & relations                       & $ 35 $ \\
                                    & triples                         & $ 4308799 $ \\
             \cmidrule(r){2-3} 
                                    & \multirow{5}{*}{data source}    & TCM \citep{yao2018topic} \\
                                    &                                 & ZzzTCM \\
                                    &                                 & Chinese Medicine Knowledge Base Website \tnote{a} \\
                                    &                                 & Chinese Medicine Family Website \tnote{b} \\
                                    &                                 & Seeking Medicine Help Website \tnote{c} \\
			\bottomrule
	\end{tabular}
    \begin{tablenotes}
        \footnotesize
        \item[a] \url{http://tcm.med.wangfangdata.com.cn}
        \item[b] \url{http://tcm.med.wangfangdata.com.cn}
        \item[c] \url{https://www.zysi.com.cn/zhongyaocai/index.html}
    \end{tablenotes}
    \end{threeparttable}}
\end{table}

\begin{figure}[t]
	\centering
	\includegraphics[width=0.99\linewidth]{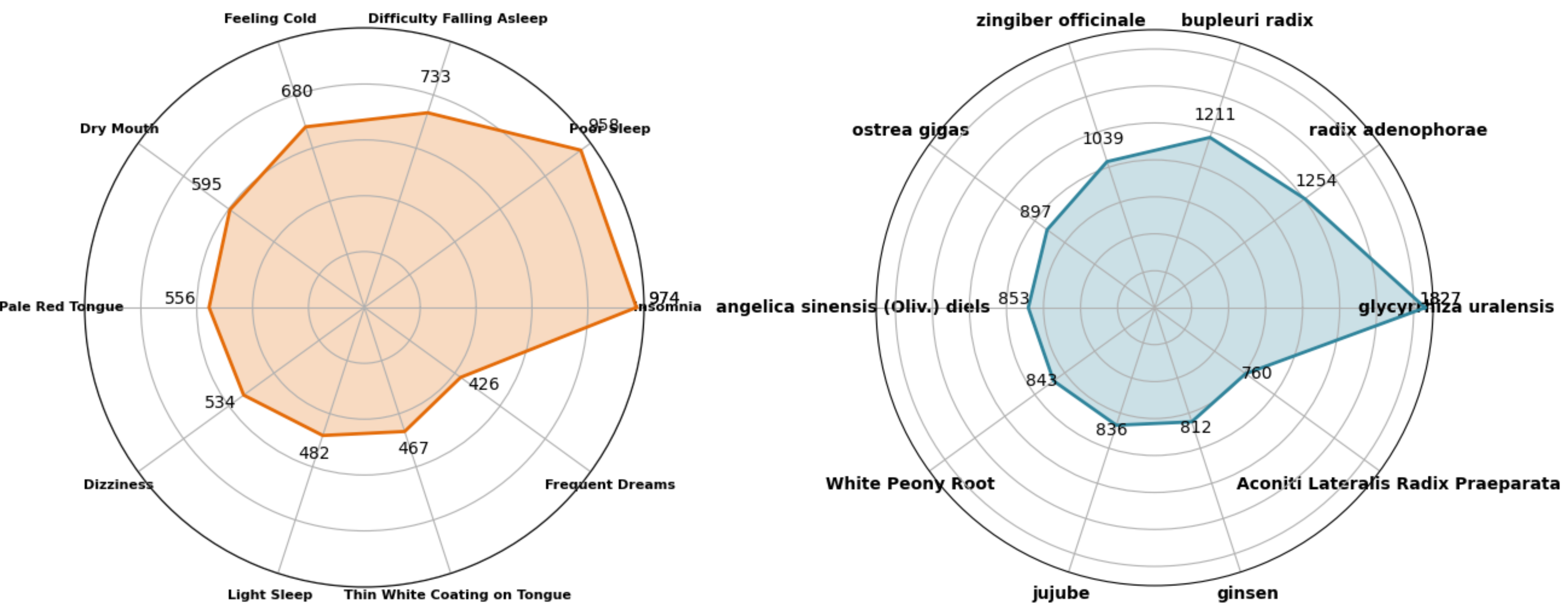}
	\caption{Visualization of Common Symptoms and Herbs Radar Chart, with orange section showing common symptoms and blue section displaying common herbs.}
	\label{fig:radar}
\end{figure}

\begin{figure}[t]
	\centering
	\includegraphics[width=0.5\linewidth]{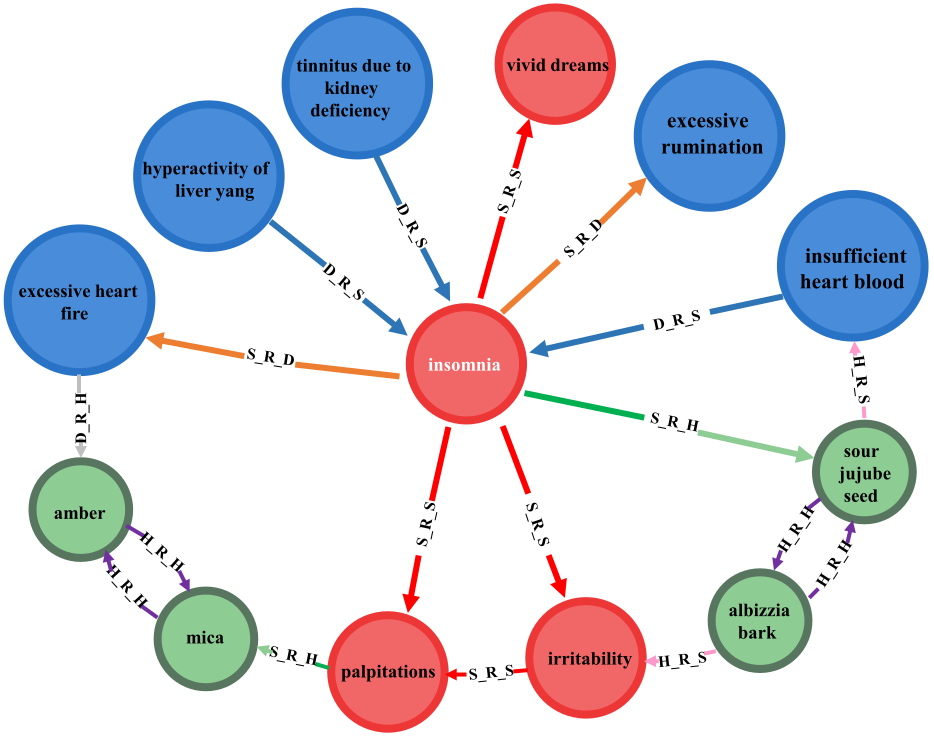}
	\caption{Local visualization of the Interaction Knowledge Graph $\mathcal{G}$. D, S, H, and R represent the disease, symptom, herb, and relationship, respectively.}
	\label{fig:IKG}
\end{figure}

\section{Additional Experiment Information}
\subsection{ZzzTCM Dataset and IKG Processing}
\label{sec:ZzzTCM Dataset and IKG Processing}
\textbf{ZzzTCM Dataset} We extracted patient narratives from the dataset provided by the Provincial Traditional Chinese Medicine (TCM) institution, focusing on symptoms related to insomnia. By merging records with the same patient EMPI and diagnosis ID, we obtained the medical histories of patients who had multiple diagnoses. During data preprocessing, we filtered out blank medical records and used the ChatGPT API called the gpt-3.5-turbo model to extract the patients' symptoms set based on the prompt "Please extract the keywords of the patient's relevant symptoms". After that, we consolidated data from all diagnoses of the same patient and transformed symptoms and herbs into multi-hot vectors before training. 


\textbf{Interaction Knowledge Graph} We construct an Interaction Knowledge Graph (IKG) from multiple data sources. The IKG contains $\mathcal{R}$ edge relations and $V$ entities, which include bidirectional edge directions, such as symptom-related herbs, and herb-related symptoms. The entities of IKG contain herbs, symptoms, diseases, pathogeny, et al. At the same time, we indexed the herbs and symptoms, starting from 0, and constructed triples based on the co-occurrence of symptoms and herbs, which are added to the constructed knowledge graph to form an Interaction Knowledge Graph (IKG). If an entity in the constructed initial knowledge graph was not found in the herbs or symptoms, we continued indexing to ensure that all entities in the IKG had index values. We further trained the IKG using completion embedding techniques.

The process of initial knowledge graph construction was divided into two main stages: first, we dynamically acquired data from websites in the relevant domains using crawling techniques such as Python, the relevant URLs are labeled in Table \ref{tab:data statistics}, which were subsequently systematically cleaned and organized. Next, the second step is to extract relevant ternary knowledge from the web data by applying manually designed rules. For example, we can convert information such as "Yang Er Ju has the efficacy of moving qi and relieving pain, and can treat wind-heat and colds" into the representation of TCM knowledge such as (Yang Er Ju, drug main treatment, wind-heat and colds) and (Yang Er Ju, drug-related effects, moving qi and relieving pain). Through these steps, we constructed the initial knowledge graph. 

The statistics of the ZzzTCM dataset and  IKG are reported in Table \ref{tab:data statistics}. In addition, common symptoms and herbal statistics are shown in Fig.\ref{fig:radar} and part of the IKG is shown in Fig.\ref{fig:IKG}.

\subsection{Metrics Details}
Given a symptom set $\mathbb{S}$ and record $r$, our proposed model generates a herb set $\hat{Y}$.
To evaluate the performance of $Top$-$K$ recommended herbs, we adopt three evaluation metrics: Precision@$Top$-$K$, Recall@$Top$-$K$, and F1@$Top$-$K$. The $Prescision$ score indicates the hit ratio of herbs is true herbs. And the $Recall$ describes the coverage of true herbs as a result of a recommendation. $F_1$ is the harmonic mean of precision and recall. In particular, we obtain the evaluation score in the test data by taking the average of patients' diagnoses.
\begin{equation}
\label{eq: Precision.}
{Precision}_{j}^{(t)}=\frac{1}{X}\frac{1}{T}\sum\limits_{j=1}^{X}\sum\limits_{t=1}^{T}\frac{| \{i: {hc}_{j, i}^{(t)}=1 \} \cap \{i: \mathrm{Top}(\hat{Y}_{j,i}^{(t)})\} |}{|\{i: \mathrm{Top}(\hat{Y}_{j,i}^{(t)})\} |}
\end{equation}

\begin{equation}
\label{eq: Recall.}
{Recall}_{j}^{(t)}=\frac{1}{X}\frac{1}{T}\sum\limits_{j=1}^{X}\sum\limits_{t=1}^{T}\frac{| \{i: {hc}_{j, i}^{(t)}=1 \} \cap \{i: \mathrm{Top}(\hat{Y}_{j,i}^{(t)})\} |}{ | \{i: {hc}_{j, i}^{(t)}=1 \} |}
\end{equation}

\begin{equation}
\label{eq: F1.}
\begin{array}{c}
\mathrm{F} 1_{j}^{(t)}=\frac{2}{\frac{1}{\operatorname{Precision}_{j}^{(t)}}+\frac{1}{\operatorname{Recall~}_{j}^{(t)}}} .
\end{array}
\end{equation}

where $hc_j^{(t)}$ is the ground-truth herb prescription during the $t$-th diagnosis of patient $j$, and $hc_{j,i}^{(t)}$ is the $i$-th element. $\mathrm{Top}(\hat{Y}_{j,i}^{(t)})$ is the top $i$-th element with the highest prediction scores. $|\cdot|$ denotes the cardinality, $\cap$ is set interaction operation. The $Prescision$ score indicates the hit ratio of herbs is true herbs. And the $Recall$ describes the coverage of true herbs as a result of a recommendation. $F_1$ is the harmonic mean of precision and recall. In particular, we obtain the evaluation score in the test data by taking the average of patients' diagnoses.

\subsection{Experimental Settings}
\label{sec:experimental setttings}
In this paper, we ran all experiments on a platform with Ubuntu 16.04 on 256GB of memory and an NVIDIA GeForce GTX 1080 Ti GPU. And we implement our model in PyTorch. In the training, we used a random seed of fixed size 2019 to guarantee the reproducibility of the results. The overall framework was optimized with Adam optimizer, where the batch size is one patient with all diagnoses and the batch size of IKG is fixed at 2048. The length $\Gamma=128$ of the record sequence $r^{(t)}$. We set the multi-hop $k$ of GNN with Hierarchical Attention-based UPDATE to three with hidden dimensions $D_k=[64,32,16]$, in order to model the third-order connectivity; The embedding size of entity $e_h$ and $e_r$ is fixed to 64. In the transition condition module, we set the hidden size $hidden\_dim = 64$ of LSTM. And we trained the model for 1000 epochs with a learning rate $lr=0.0001$ and the coefficient of normalization $\lambda_\Theta =10^{-5}$. For evaluation metrics, we set $Top$-$K = [5,10,20]$. We report the average metrics for all patients in the test set. Moreover, an early stopping strategy is suggested \citep{wang2019kgat}, premature stopping if $recall@Top$-$K=20$ on the validation set does not increase for $early\_stop=50$ successive epochs. The default Xavier initializer \citep{glorot2010understanding} to initialize the model parameters. Also, we conduct experiments on parameter sensitivity, which are presented in Appendix \ref{sec:parameter sensitivity}.

\subsection{Baselines Details}
In this paper, we evaluate the performance of SCEIKG by comparing it against the following baselines. To carry out a fair comparison, all experiments are run on the same platform. We also utilize 64 batch sizes for traditional recommendation approaches and one patient for sequence-based models. Also, the early stop mechanism is also applied in the baseline methods and the number of layers is set to 3 for GCN-based models.

\textbf{BPR} \citep{rendle2012bpr} performs poorly due to its neglect of multi-hop interactions and the evolving nature of a patient's condition. It presents a generic optimization criterion BPR-Opt for personalized ranking which is the maximum posterior estimator derived from a Bayesian analysis of the traditional recommendation. 64-dim embedding tables implement the model, the learning rate $10^{-3}$, 64 batch size, and the Adam as optimizer. Also, we utilize an early stop mechanism to train all models.

\textbf{GCN} \citep{kipf2016semi} introduces the degree matrix of the node to solve the problem of self-loops and the normalization of the adjacency matrix and sums the embedding of neighbor nodes to update the current node. The parameter settings of GCN are the same as the SMGCN \citep{jin2020syndrome}.

\textbf{KGAT} \citep{wang2019kgat} incorporates higher-order collaborative signals for traditional recommendations, it falls short in exploring the higher-order relationships specific to TCM recommendations, namely the connections between symptom sets and herb sets. Following the original paper, we implement the 64-dim embedding tables. Adam is used as the optimizer with a learning rate at $10^{-4}$.

\textbf{GAMENet} \citep{shang2019gamenet} and \textbf{SafeDrug} \citep{yang2021safedrug} capture comprehensive medical histories of patients utilizing longitudinal vectors of medical codes. These models solely consider the patient's medication records and fail to capture the nuanced aspects of the physique. Although GAMENet exhibits some similarities to our model when $Top$-$K$ is set to 20, its performance lags behind ours for other $Top$-$K$ values, thus underscoring the strength of our patient history-based approach. We use the same suit of hyperparameters reported in the original papers: the learning rate at $5 \times 10^{-4}$ use 64-dim embedding tables and 64-dim GRU as RNN.

\textbf{SMGCN} \citep{jin2020syndrome} obtains the embedding of symptoms and herbs and recommends herbs through an implicit syndrome induction process. For SMGCN, the learning rate is $2e-4$ and the dimension of the GCN layer is 128. The regularization coefficient is set to $7 \times 10^{-3}$, the dimensions of the embedded layer and the hidden layer are 64, the GCN output dimension of the last layer is 256 and the MLP layer size is 256.

\textbf{KDHR} \citep{yang2022multi} introduces herb properties as additional auxiliary information by constructing an herb knowledge graph and employs a graph convolution model with multi-layer information fusion to obtain symptom and herb feature representations. the initial learning rate is $3 \times 10^{-4}$, Adam is used to optimize the parameters, and the regularization coefficient is set to 0.007.

Although \textbf{SMGCN} \citep{jin2020syndrome} and \textbf{KDHR} \citep{yang2022multi} achieve excellent accuracy for TCM recommendation, both models neglect the crucial aspect of accounting for changes in a patient's condition over time. Note that GAMENet and SafeDrug are not considered baseline since they require extra ontology data. Also, when applying our dataset in KDHR, we removed the module for the herb knowledge graph. 

\section{Training Algorithm for SCEIKG model and Inference}
\label{sec: Training Algorithm for SCEIKG model and Inference}
\begin{algorithm}[H]
    \caption{Training of SCEIKG}
    \label{algo:SCEIKG}
    \begin{algorithmic}
        \REQUIRE Training set $\iota$, Interaction knowledge graph $\mathcal{G}$, weight matrix of IKG $\mathbb{W}$ in Eq.(1), batch of patients $\zeta$, batch of Triplets $\xi$, total number of patients $\eta$, total number of epoch $E$, the configuration $\Theta$
        \STATE \textbf{Output: herb set $\hat{Y}$} 
        \STATE Initialize all configurations $\Theta$
        \FOR {epoch $\leftarrow 0,1,\cdot \cdot \cdot, $E$ $}
        \STATE Generate Entities Embedding $E \in \mathbb{R}^{V \times D_{out}}$, propagate over the interaction knowledge graph
        \STATE /*Phase I: Interaction Knowledge graph Complementation*/
            \FOR {triples $(h,r,t)$ in $\mathcal{G}$ of batch $\xi$ }
            \STATE Calculate the score of the knowledge triples $\boldsymbol{f}(h,r,t)$
            \STATE Calculate the interaction knowledge graph loss $L_{IKG}$ and update interaction knowledge graph embedding $e_h$.
            \ENDFOR
        \STATE Update the weight matrix $\mathbb{W}$ of $\mathcal{G}$ by the function $\mathrm{UPDATE}(\cdot)$
        \STATE /*Phase II: Recommended herbs based on sequential diagnoses for each patient*/
            \FOR{batch j:=1 to $\frac{\eta}{\zeta}$}
            \STATE Select the batch of patients sequential records $ \Omega$
            \STATE /*Note that the current diagnosis contains $\xi$ patients*/
                \FOR{diagnosis $t:=1$ to $|T|$}
                \IF{t==1}
                \STATE Select the $t$-th batch of patient, $\Omega^{(t)}$
                \ELSE 
                \STATE Select the $t$-th batch of patient, $\Omega^{(t)}$ and Transition condition representation $\Psi^{(t-1)}$
                \ENDIF
                \STATE Generate Condition Representation $h^{(t)}_\mathbb{C}$ and Symptom Representation $h^{(t)}_\mathbb{S}$ 
                \STATE /*The first diagnosis is not having the previous patient's condition and the last diagnosis is not having the next condition*/
                \STATE Generate Horizontal Patient Representation $\Phi^{(t)}_P$ based on the Transition Condition Module $\Psi^{(t-1)}$
                \STATE Generate Patient to herb Matching ${\hat{Y}}^{(t)}$
                \ENDFOR
            \STATE Accumulate the loss of herb prediction and update the configuration $\Theta$ by Adam
            \ENDFOR
        \ENDFOR
    \end{algorithmic}
\end{algorithm}

\begin{table}[t]
    \centering
    \caption{Hyperparameter experiment results.}
    \label{tab:hyperparameter}
    \resizebox{\textwidth}{!}{
	\begin{tabular}{c c c c c c c c c c c} 
	      \toprule
            \multicolumn{2}{c}{\multirow{2}{*}{\textbf{Hyperparameters}}} & \multicolumn{3}{c}{\textbf{Precision}} & \multicolumn{3}{c}{\textbf{Recall}} & \multicolumn{3}{c}{\textbf{F1}}\\ 
            \cmidrule(r){3-5} \cmidrule(r){6-8} \cmidrule(r){9-11}
			& & P@5 & P@10 & P@20 & R@5 & R@10 & R@20 & F1@5 & F1@10 & F1@20   \\
            \midrule
			\multirow{4}{*}{$lr$}  & 0.01      & $0.1664$ & $0.0089$ & $0.0758$ & $0.0885$ & $0.0938$ & $0.1544$ & $0.1096$ & $0.0872$ & $0.0984$\\
            & 0.001      & $0.4322$ & $0.3490$ & $0.2638$ &$0.1953$ & $0.3227$ & $0.5000$ & $0.2633$ & $0.3249$ & $0.3361$\\
            & 0.0001$^*$      & $\textbf{0.5477}$ & $\textbf{0.4275}$ & $\textbf{0.3087}$ & $0.2727$ & $0.4243$ & $\textbf{0.6010}$ & $0.3538$ & $0.4128$ & $\textbf{0.3973}$\\
			& 0.00001     & $0.5383$ & $0.4248$ & $0.2990$ & $\textbf{0.2807}$ & $\textbf{0.4318}$ & $0.5947$ & $\textbf{0.3550}$ & $\textbf{0.4130}$ & $0.3867$\\
            \midrule
            \multirow{3}{*}{$\lambda_{\Theta}$} & $1.0\times10^{-4}$   & $0.5302$ & $0.4101$ & $\textbf{0.2883}$ & $0.2688$ & $0.4091$ & $\textbf{0.5655}$ & $0.3466$ & $0.3973$ & $\textbf{0.3721}$\\
            & $1.0\times10^{-5}$ $^*$ & $\textbf{0.5477}$ & $\textbf{0.4275}$ & $\textbf{0.3087}$ & $\textbf{0.2727}$ & $0.4243$ & $\textbf{0.6010}$ & $\textbf{0.3538}$ & $0.4128$ & $\textbf{0.3973}$\\
            & $1.0\times10^{-6}$    & $0.5436$ & $0.4349$ & $0.3027$ & $0.2731$ & $0.4337$ & $0.5902$ & $0.3534$ & $0.4209$ & $0.3896$\\
            \midrule
            \multirow{3}{*}{$\Gamma$} & 32   & $0.5409$ & $0.4376$ & $0.3047$ & $0.2724$ & $\textbf{0.4355}$ & $0.5914$ & $0.3521$ & $\textbf{0.4232}$ & $0.3916$\\
            & 64 & $0.5208$ & $0.4060$ & $0.2849$ & $0.2654$ & $0.4200$ & $0.5748$ & $0.3418$ & $0.3982$ & $0.3700$\\
            & 128$^*$ & $\textbf{0.5477}$ & $\textbf{0.4275}$ & $\textbf{0.3087}$ & $\textbf{0.2727}$ & $0.4243$ & $\textbf{0.6010}$ & $\textbf{0.3538}$ & $0.4128$ & $\textbf{0.3973}$\\
            \midrule
            \multirow{3}{*}{$hidden\_{dim}$} & 32   & $0.5423$ & $\textbf{0.4403}$ & $0.3007$ & $0.2692$ & $\textbf{0.4314}$ & $0.5829$ & $0.3496$ & $\textbf{0.4225}$ & $0.3864$\\
            & 64$^*$ & $\textbf{0.5477}$ & $0.4275$ & $\textbf{0.3087}$ & $\textbf{0.2727}$ & $0.4243$ & $\textbf{0.6010}$ & $\textbf{0.3538}$ & $0.4128$ & $\textbf{0.3973}$\\
            & 128    & $0.5289$ & $0.4262$ & $0.3070$ & $0.2687$ & $0.4281$ & $0.5979$ & $0.3464$ & $0.4142$ & $0.3954$\\
             & 256    & $0.5289$ & $0.4208$ & $0.3007$ & $0.2682$ & $0.4261$ & $0.5940$ & $0.3455$ & $0.4083$ & $0.3881$\\
			\bottomrule
		\end{tabular}}
\begin{tablenotes}
        \footnotesize
        \item * Asterisks indicate baseline experiment settings
\end{tablenotes}
\end{table}

\textbf{Algorithm} We provide further insights into the implementation of our Algorithm \ref{algo:SCEIKG}, which is rooted in the Expectation Maximization (EM) algorithm \citep{dempster1977maximum}, a well-established iterative optimization strategy. Our model is built upon a similar conceptual framework as the EM algorithm. Initially, we engage in complementary learning of the knowledge graph, involving updates to the Interaction Knowledge Graph (IKG) to enhance the embedded representations of entities. Subsequently, these enriched entity embeddings are applied in the training phase for TCM recommendations. These two components of the cycle iteratively interact until the model converges, and the optimization process concludes.

As illustrated in Fig.\ref{fig:figure2}, the IKG updates play a pivotal role in refining the individual modules depicted in the figure. Conversely, the training of the model reciprocally enhances the IKG, as shown on the right. This dynamic interaction fosters iterative improvement. It is well-recognized that constructing a comprehensive knowledge graph for TCM is an intricate task that necessitates extensive data support. Therefore, knowledge graph complementation, which involves inferring new information from existing data. In the first phase of Algorithm \ref{algo:SCEIKG} \textbf{(Phase I: Interaction Knowledge graph Complementation)}. The representation entities in the IKG are learned by Eq.\ref{eq: equation3} and then updated inversely by $L_{IKG}$ in the loss function in Eq.\ref{eq: equation6}, which complements and enriches the information of the entities. The backpropagation process of the first phase is shown,

\begin{equation}
\begin{aligned}
\frac{\partial \mathcal{L}_{I K G}}{\partial \boldsymbol{f}\left(h, r, t^{\prime}\right)} &=-\frac{1}{\sigma\left(\boldsymbol{f}\left(h, r, t^{\prime}\right)-\boldsymbol{f}(h, r, t)\right)} \cdot \frac{d}{d x} \sigma\left(\boldsymbol{f}\left(h, r, t^{\prime}\right)-\boldsymbol{f}(h, r, t)\right) \\
&=-\frac{1}{\sigma\left(\boldsymbol{f}\left(h, r, t^{\prime}\right)-\boldsymbol{f}(h, r, t)\right)} \cdot \sigma\left(\boldsymbol{f}\left(h, r, t^{\prime}\right)-\boldsymbol{f}(h, r, t)\right) \cdot\left(1-\sigma\left(\boldsymbol{f}\left(h, r, t^{\prime}\right)-\boldsymbol{f}(h, r, t)\right)\right) \\
&=-\left(1-\sigma\left(\boldsymbol{f}\left(h, r, t^{\prime}\right)-\boldsymbol{f}(h, r, t)\right)\right)
\end{aligned}   
\end{equation}

\begin{equation}
\begin{aligned}
\frac{\partial \mathcal{L}_{I K G}}{\partial \boldsymbol{f}(h, r, t)} &=\frac{1}{\sigma\left(\boldsymbol{f}\left(h, r, t^{\prime}\right)-\boldsymbol{f}(h, r, t)\right)} \cdot \frac{d}{d x} \sigma\left(\boldsymbol{f}\left(h, r, t^{\prime}\right)-\boldsymbol{f}(h, r, t)\right) \\
&=\frac{1}{\sigma\left(\boldsymbol{f}\left(h, r, t^{\prime}\right)-\boldsymbol{f}(h, r, t)\right)} \cdot \sigma\left(\boldsymbol{f}\left(h, r, t^{\prime}\right)-\boldsymbol{f}(h, r, t)\right) \cdot\left(1-\sigma\left(\boldsymbol{f}\left(h, r, t^{\prime}\right)-\boldsymbol{f}(h, r, t)\right)\right) \\
&=1-\sigma\left(\boldsymbol{f}\left(h, r, t^{\prime}\right)-\boldsymbol{f}(h, r, t)\right)
\end{aligned}  
\end{equation}

the gradient descent derivation for $L_{IKG}=\sum\limits_{\left(h,r,t,t'\right) \in \mathcal{T}} -ln\sigma\left(\boldsymbol{f}\left(h,r,t'\right)-\boldsymbol{f}\left(h,r,t\right)\right)$. Then, we compute the gradient of the loss with respect to  $\boldsymbol{f}\left(h, r, t^{\prime}\right)$. Similarly, compute the gradient of the loss with respect to  $\boldsymbol{f}(h, r, t)$. Then, compute the gradients of  $\boldsymbol{f}\left(h, r, t^{\prime}\right)$  and  $\boldsymbol{f}(h, r, t)$  with respect to their respective embeddings:

\begin{equation}
\begin{array}{l}
\frac{\partial \boldsymbol{f}\left(h, r, t^{\prime}\right)}{\partial h}=\frac{\partial}{\partial h}\left(C \cdot\left\|\sin \left(W_{r}e_h+e_{r}-W_{r}e_{t^{\prime}}\right)\right\|_{1}\right)=C
\frac{\partial}{\partial h}\left\|\sin \left(W_{r}e_h+e_{r}-W_{r}e_{t^{\prime}}\right)\right\|_{1} \\
\end{array}
\end{equation}

\begin{equation}
\begin{array}{l}
\frac{\partial \boldsymbol{f}(h, r, t)}{\partial h}=\frac{\partial}{\partial h}\left(C \cdot\left\|\sin \left(W_{r}e_h+e_{r}-W_{r}e _t\right)\right\|_{1}\right)=C
\frac{\partial}{\partial h}\left\|\sin \left(W_{r}e_h+e_{r}-W_{r}e_t\right)\right\|_{1}
\end{array}
\end{equation}

In gradient descent, update the parameters in the opposite direction of the gradient to minimize the loss function. Assuming that the parameters influencing  $f(h, r, t)$  are denoted by  $\theta$. The update rule for the parameters would be,

\begin{equation}
\begin{array}{l}
\theta \leftarrow \theta-\eta \cdot \frac{\partial L_{I K G}}{\partial \boldsymbol{f}(h, r, t)} \cdot \frac{\partial \boldsymbol{f}(h, r, t)}{\partial \theta}
\end{array}
\end{equation}
Where $\eta$  is the learning rate, $\frac{\partial L_{I K G}}{\partial \boldsymbol{f}(h, r, t)}$  is the gradient we calculated, $\frac{\partial \boldsymbol{f}(h, r, t)}{\partial \theta}$  is the gradient of the score function with respect to the parameters $\theta$. Here, we provide an approximate derivation of the derivation descent derivation equation for $L_{IKG}$.

In traditional knowledge graph embedding methods only individual knowledge triples can be represented efficiently, and higher-order similarities between entities cannot be captured. However, this higher-order information is critical for understanding complex interactions between entities, especially in the field of herbal medicine recommendation. We detail how the entity representations obtained in Phase I can be applied to the recommendation of herbs in Phase II \textbf{(Phase II: Recommended herbs based on sequential diagnoses for each patient)} using a Graph Neural Network (GNN) approach. In this process, we use the adjacency matrix, which is constructed by the normalized pointwise mutual information method while utilizing the entity knowledge representations learned in the first stage. Then, we use the following matrix according to Eq.\ref{eq: equation2} to aggregate higher-order relationships, as well as similarity information between higher-order entities. In addition, we introduce an attention mechanism for entity correlation, which is used to update the structure of the IKG graph, thus further enriching the information on herbal recommendations. Ultimately, we backpropagate through the loss function in Eq.\ref{eq: equation6}. This loss function includes a mean square error loss $L_{mse}$ for measuring the distance between the recommended set of herbs and the actual set of herbs, and a state loss $L_{state}$ for measuring the similarity of the states before and after the recommendation of the herbs as well as a regularization term $\boldsymbol{R}$ for placing constraints on the relationships between herbs. The specific back propagation derivation is shown below,

\begin{equation}
\frac{\partial L_{m s e}}{\partial \Theta}=2 \sum_{i=1}^{N}\left(hc_{i}^{(t)}-\hat{Y}_{i}^{(t)}\right) \cdot \frac{\partial \hat{Y}_{i} ^{(t)}}{\partial \Theta}
\end{equation}

Then, calculate the gradient of $L_{\text {state }}$  with respect to the parameter $\Theta$,
\begin{equation}
\frac{\partial L_{s t a t e}}{\partial \Theta}=\frac{\partial}{\partial \Theta}\left(\frac{h_\mathbb{C}^{(t+1)} \cdot \Psi^{(t)}}{\left\|h_\mathbb{C}^{(t+1)}\right\| \cdot\left\|\Psi^{(t)}\right\|}\right)  
\end{equation}

Since $h_\mathbb{C}^{(t+1)}$ and $\Psi^{(t)}$ have nothing to do with the parameter $\Theta$, we only need to calculate the gradient of the numerator part. Finally, the gradient of the regularisation term $\boldsymbol{R}$ with respect to the parameter $\Theta$ is calculated as follows.

\begin{equation}
\frac{\partial \boldsymbol{R}}{\partial \Theta}=-\sum_{i=1}^{N} \sum_{j=1}^{N} \mathbb{W}_{i j}^{h} \hat{Y}_{i}^{(t)} \cdot \frac{\partial \hat{Y}_{i}^{(t)}}{\partial \Theta} \cdot \hat{Y}_{j}^{(t)}-W_{i j}^{h} \hat{Y}_{i}^{(t)} \cdot \hat{Y}_{j}^{(t)} \cdot \frac{\partial \hat{Y} _{j}^{(t)}}{\partial \Theta}
\end{equation}

\begin{figure}[t]
	\centering
	\includegraphics[width=0.99\linewidth]{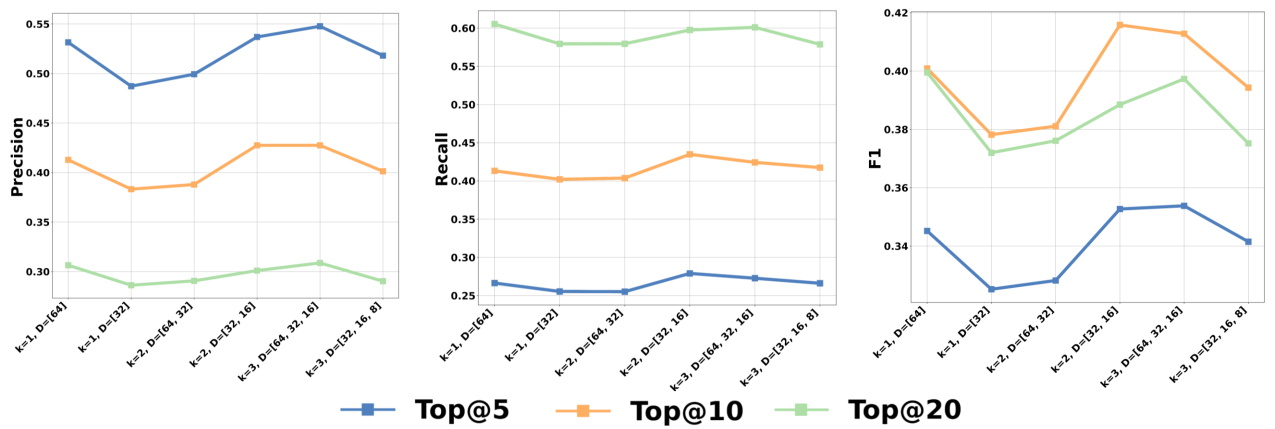}
	\caption{Effect of dimensions with different layers on SCEIKG.}
	\label{fig:dimensions}
\end{figure}

\begin{table}[t]
    \centering
    \caption{The difference and intersection herbs prescribed by SCEIKG and w/o IKG according to clinical symptoms and records of the same patient for two diagnoses.}
    \label{tab: herb example2}
    \resizebox{\textwidth}{!}{
	\begin{tabular}{c l l l} 
	      \toprule
            \makecell[c]{\multirow{2}{*}{\textbf{Sequential diagnoses}}}  &\makecell[l]{\multirow{2}{*}{\textbf{Symptom Set}}} & \multicolumn{2}{c}{\textbf{Herb Set}} \\ 
            \cmidrule(r){3-4} 
			&  & \makecell[c]{\textbf{w/o IKG}} & \makecell[c]{\textbf{SCEIKG}} \\
            \midrule

             & 抑郁症(depression)  & \red{\textbf{黄芩(scutellaria baicalensis)}} & \red{\textbf{黄芩(scutellaria baicalensis)}} \\
             & 口干(xerostomia)  & \red{\textbf{炙甘草(glycyrrhiza uralensis)}} & \red{\textbf{炙甘草(glycyrrhiza uralensis)}} \\
			& 大便费力(dyschezia)  & \red{\textbf{生姜(ginger)}} & \red{\textbf{生姜(ginger)}} \\
            & 入睡困难(insomnia)  & \red{\textbf{大枣(jujube)}} & \red{\textbf{大枣(jujube)}} \\
First diagnosis   & 眠浅易醒(light sleep, easy to wake up)  & \red{\textbf{人参(ginsen)}} & \red{\textbf{人参(ginsen)}} \\
            & 乏力(fatigue)  & 桂枝(cinnamomum cassia) & 桂枝(radix adenophorae) \\
            & 胸闷(chest tightness)& 茯苓(tuckahoe) & 茯苓(bupleuri radix) \\
            & 四肢麻木(numbness of limbs)  & \blue{\textbf{川芎(sichuan lovage rhizome)}} & \blue{\textbf{白芍(paeonia lactiflora)}} \\
            & 舌淡红(pale red tongue)  &\blue{\textbf{法半夏(pinellia tuber)}}  & \blue{\textbf{牡蛎(ostrea gigas)}} \\
            & 下睑淡白(pale lower eyelid)  & \blue{\textbf{当归(angelica sinensis (Oliv.) diels)}} & \blue{\textbf{干姜(zingiber officinale)}} \\
            \cmidrule(r){1-4}
            \multicolumn{4}{c}{\textbf{p@10=0.5000 r@10=0.6250 f1@10=0.5556}} \\
            \cmidrule(r){1-4}
            & 口干(xerostomia)  & \red{\textbf{黄芩(scutellaria baicalensis)}} & \red{\textbf{黄芩(scutellaria baicalensis)}} \\
            & 惊恐(panic)  & \red{\textbf{赤芍(paeonia lactiflora)}} & \red{\textbf{赤芍(paeonia lactiflora)}} \\
            & 焦虑 (anxiety)  & \red{\textbf{炙甘草(glycyrrhiza uralensis)}} & \red{\textbf{炙甘草(glycyrrhiza uralensis)}} \\
            & 入睡困难(insomnia)  & \red{\textbf{大枣(jujube)}} & \red{\textbf{大枣(jujube)}} \\
            & 眠浅易醒(light sleep, easy to wake up)  & \red{\textbf{生姜(ginger)}} & \red{\textbf{生姜(ginger)}} \\
            & 乏力(fatigue)  & 茯苓(tuckahoe) & \red{\textbf{清半夏(pinellia tuber)}} \\
Second diagnosis    &  胸闷(chest tightness)  & 人参(ginsen) & 人参(ginsen)  \\
            &  四肢麻木(numbness of limbs)  & 桂枝(cinnamomum cassia) & 桂枝(cinnamomum cassia)  \\
            &  小便频急(frequent urination)  & \blue{\textbf{甘草(glycyrrhiza uralensis fisch)}} & 茯苓(bupleuri radix) \\
            &  右手心热(palm heat)  & \blue{\textbf{当归(angelica sinensis (Oliv.) diels)}}  & \blue{\textbf{炒六神曲(medicated leaven)}} \\
            &  舌淡红(pale red tongue)  &  &  \\
            &  苔薄(thin fur)  &  &  \\
            &  下睑淡白边偏红 (pale lower eyelid with reddish edges)  &  & \\
            \cmidrule(r){1-4}
            \multicolumn{4}{c}{\textbf{p@10=0.5000 r@10=0.8333 f1@10=0.6250}} \\
			\bottomrule
		\end{tabular}}
\end{table}

Combining the above three components, the individual gradients are summed up to give the total gradient $\frac{\partial}{\partial \Theta}\left(L_{m s e}+L_{s t a t e}+\lambda \boldsymbol{R}\right)$. This total gradient will be used in the gradient descent optimization algorithm to update the parameter $\Theta$ to minimize the overall loss function. Ultimately, by continually iterating this gradient descent process, we can optimize the parameters of the model, thus minimizing the overall loss function and achieving the optimization goal of our model. These two stages iteratively update each other and finally complete the training of the whole model.

Note that for each patient at the first diagnosis, we do not have access to the patient's previous state and therefore cannot perform state transfer prediction. Only after the first diagnosis can we start modeling the patient's historical state dynamically. At the last diagnosis, we did not acquire the patient's next state either, so we need to pay attention to the boundary condition handling in modeling. In addition, there are different numbers of diagnoses for each patient, so we first pad the batch of patients to the same number of diagnoses, but during training, the padding data is not entered into forward propagation. Hence, we can recommend multiple patients in parallel.

\textbf{Training Inference} The model is trained end-to-end. We optimize the prediction loss and $L_{IKG}$ alternatively. In particular, for a batch of randomly sampled $(h,r,t,t')$, we update the embeddings for all nodes; hereafter, we sample a batch of patients with consecutive diagnoses randomly, retrieve their representations after $L$ steps of propagation, and then update model parameters by utilizing the gradients of the prediction loss. Finally, we select the top $K$ herbs with the highest probabilities as the recommended herb set.

\begin{figure}[t]
	\centering
	\includegraphics[width=0.99\linewidth]{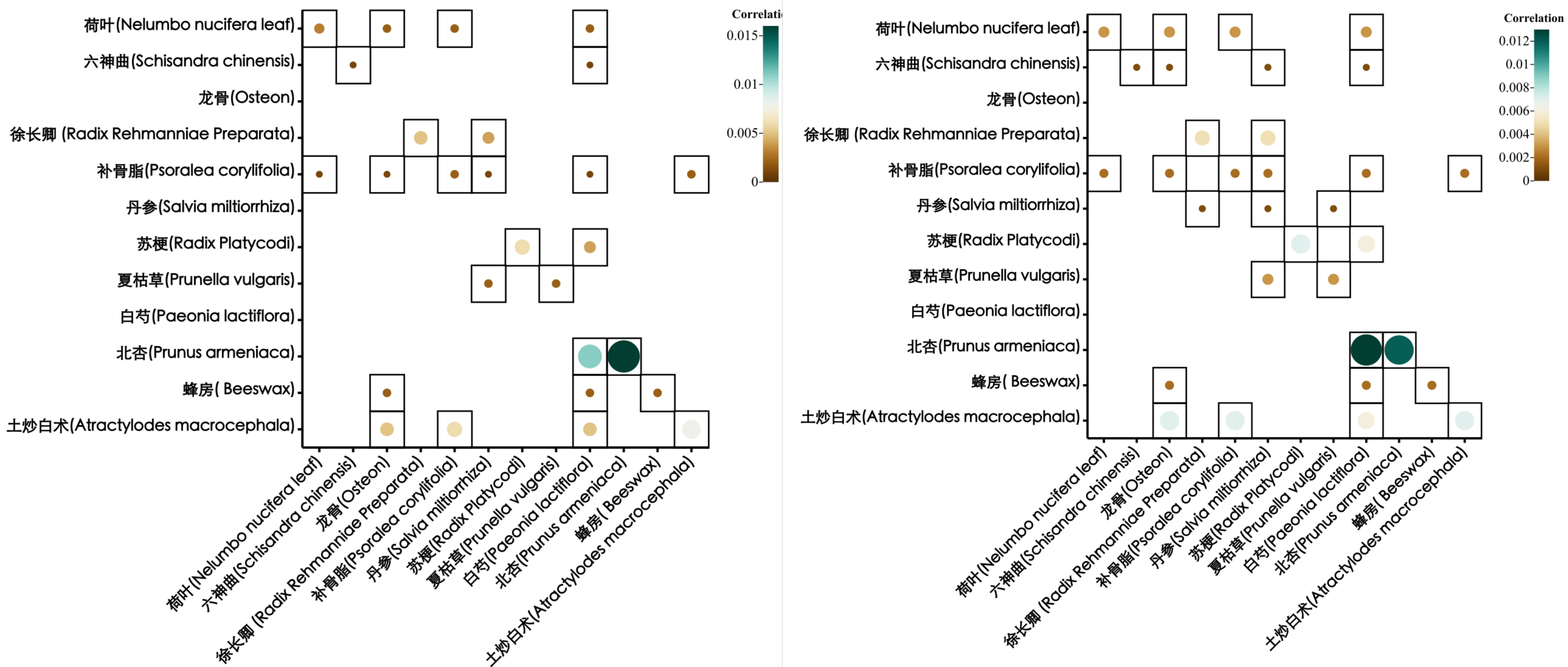}
	\caption{The visualization of the heatmaps on the relationship between partly herbal pairs. (a) Partly herb pairs, derived from a constructed weight matrix that captures the co-occurrence of all entities involved; (b) Partly herb pairs, through the training of our SCEIKG model.}
	\label{fig:figure4}
\end{figure}

\begin{figure}[t]
	\centering
	\includegraphics[width=0.5\linewidth]{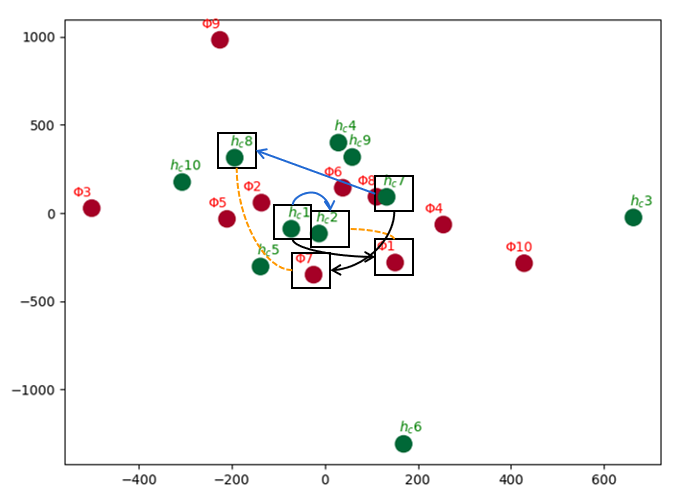}
	\caption{Conditional embedding visualization. The black solid line represents the transformation of the patient's current diagnosis status to the status after taking medication, and the blue solid line indicates the transition from the current diagnosis status to the status at the next diagnosis. On the other hand, the orange dashed line represents the extent to which the current status after taking medication transfers to the status at the next diagnosis.}
	\label{fig:condition_emb}
\end{figure}


\section{Parameter Sensitivity}
\label{sec:parameter sensitivity}
In this section, we apply a grid search for hyper-parameters: the learning rate is tuned amongst $lr=\{0.01,0.001,0.0001,0.00001\}$, the coefficient of normalization $\lambda_{\Theta}$ is searched in $\{10^{-4},10^{-5},10^{-6}\}$. We tune the max length $\Gamma =\{32, 64, 128\}$ of the condition to explore the impact of changes in historical medication patient status. The hidden dimensional size of LSTM $hidden\_dim=\{32,64,128\}$ to capture the useful information across multiple diagnoses. Hyperparameter experiment results are provided in Table \ref{tab:hyperparameter}. The model tends to be robust to hyperparameter changes. Also, we explore whether our proposed model can benefit from a larger number of embedding propagation layers, we tune the depth of GNN layers on the submodel, which is varied in $k=\{1, 2, 3\}$ combined with the different dimensions $kd$ of each layer. The result is shown in Fig.\ref{fig:dimensions}. Intuitively, this is because more vectors can encode more useful information in latent space. However, due to the limitations of large knowledge graphs received from experimental conditions, we are not able to conduct higher dimensional exploration.

\section{Additional Case Study}

\label{sec:additional case study}


\textbf{Herb Recommendation} Table.\ref{tab: herb example2} presents the herbal recommendations of models both without and with IKG. In the table, herbal recommendations consistent with those of real TCM doctors are highlighted in red, while the blue font indicates inconsistency between the recommendations of the two models. Despite observing slight advantages for the model without IKG in some evaluation metrics during ablation experiments, it is noteworthy that, the recommendations generated by SCEIKG are more in line with classical prescriptions found in ancient records. Generally, these classic prescriptions are considered to have been validated over thousands of years and are therefore potentially more suitable. This suggests that our recommendations may be more rational or better aligned with historical usage. It also indicates that our IKG provides richer information, while models without IKG rely solely on the underlying data for recommendations. From the perspective of herbal combinations, IKG furnishes more in-depth and comprehensive information for TCM recommendations, facilitating better decision-making by medical professionals.


\textbf{Herb Compatibility} As illustrated in Fig.\ref{fig:figure4}, we unveil the shift in the correlation between partly herb pairs, which captures the co-occurrence of all entities involved. For instance, the connection between (schisandra chinensis, osteon) and (radix rehmanniae preparata, salvia miltiorrhiza). These once-disparate pairs now harmoniously coexist within the same prescription. The change reverberates as a testament to the constraints imposed on the compatibility of these herb pairs. The interpretative experiments of the embedding visualization and herb recommendations on the ZzzTCM dataset are given in the supplementary material.

\textbf{Embedding Visualization} To show a more intuitive understanding of the changes in patient condition. We utilized the t-SNE\citep{van2008visualizing} to portray the patient's real condition embeddings $h_c$ and horizontal patient condition embeddings $\Phi$. As shown in Fig.\ref{fig:condition_emb}, it becomes apparent that patients' condition changes tend to cluster together, while also allowing for some isolated instances. This intriguing phenomenon arises from the limited correlation observed between a patient's current diagnosis and their previous diagnoses. The patient's condition $h_c^1$ during their initial diagnosis. As the patient follows the prescribed medication, a state transition embedding, referred to as $\Phi^1$, occurs, and we observed a relatively small degree of state transfer from $h_c^1$ to both $\Phi^1$ and $h_c^2$. In subsequent diagnoses, we can observe a significant distance discernible between the state $\Phi^7$ after the transfer of $h_c^7$ to the predicted state and the state $h_c^8$ at the next diagnosis, with different degrees of state transfer used to assist the current true state for medication recommendation, thus improving the accuracy of the recommendation. The difference in the degree of state transfer is due to the fact that the patient will not respond to the recommended remedy to the same degree. However, we use implicit state transfer to assist in subsequent diagnoses, and in future work, we will represent our states in a more direct way.

\end{CJK}
\end{document}